\documentclass[10pt,journal,compsoc]{IEEEtran}
\usepackage{amsmath,amsfonts}
\usepackage{algorithmic}
\usepackage{algorithm}
\usepackage{array}
\usepackage[caption=false,font=normalsize,labelfont=sf,textfont=sf]{subfig}
\usepackage{textcomp}
\usepackage{stfloats}
\usepackage{url}
\usepackage{verbatim}
\usepackage{graphicx}
\usepackage{cite}
\usepackage{booktabs}
\usepackage{gensymb}
\usepackage{times}
\usepackage{epsfig}
\usepackage{amssymb}
\usepackage{amsbsy}
\usepackage{epstopdf}
\usepackage{float}
\usepackage{tabularx, booktabs}
\usepackage{multirow}
\usepackage{commath}
\usepackage{blindtext}
\usepackage{mathtools}
\usepackage{bbding}
\usepackage{makecell}
\usepackage[T1]{fontenc}
\usepackage{xcolor}

\begin{document}

\title{Advances in 3D Generation: A Survey}

\author{IEEE Publication Technology,~\IEEEmembership{Staff,~IEEE,}
\thanks{This paper was produced by the IEEE Publication Technology Group. They are in Piscataway, NJ.}
\thanks{Manuscript received April 19, 2021; revised August 16, 2021.}}

\markboth{Journal of \LaTeX\ Class Files,~Vol.~14, No.~8, August~2021}%
{Shell \MakeLowercase{\textit{et al.}}: A Sample Article Using IEEEtran.cls for IEEE Journals}

\IEEEpubid{0000--0000/00\$00.00~\copyright~2021 IEEE}

\IEEEtitleabstractindextext{%
\begin{abstract}
Generating 3D models lies at the core of computer graphics and has been the focus of decades of research. With the emergence of generative artificial intelligence (AI) and advanced generative models, the field of 3D content generation is rapidly advancing, unlocking unprecedented capabilities for creating high-quality and diverse 3D models. The rapid growth of this field makes it challenging to keep up with all recent developments. This survey aims to introduce the fundamental methodologies of 3D generation methods and establish a structured roadmap, encompassing 3D representation, generation methods, datasets, and corresponding applications. Specifically, we introduce the 3D representations that serve as the backbone for 3D generation. Furthermore, we provide a comprehensive overview of the rapidly growing literature on generation methods, categorized by the type of algorithmic paradigms, including feedforward generation, optimization-based generation, procedural generation, and generative novel view synthesis. Lastly, we discuss available datasets, applications, and open challenges. This survey offers an intuitive starting point for researchers, artists, and practitioners alike to explore this exciting topic and foster further advancements in the field of 3D content generation.

\end{abstract}

\begin{IEEEkeywords}
Article submission, IEEE, IEEEtran, journal, \LaTeX, paper, template, typesetting.
\end{IEEEkeywords}
}

\maketitle

\begin{figure*}[h]
\begin{center}
   \includegraphics[width=0.95\linewidth]{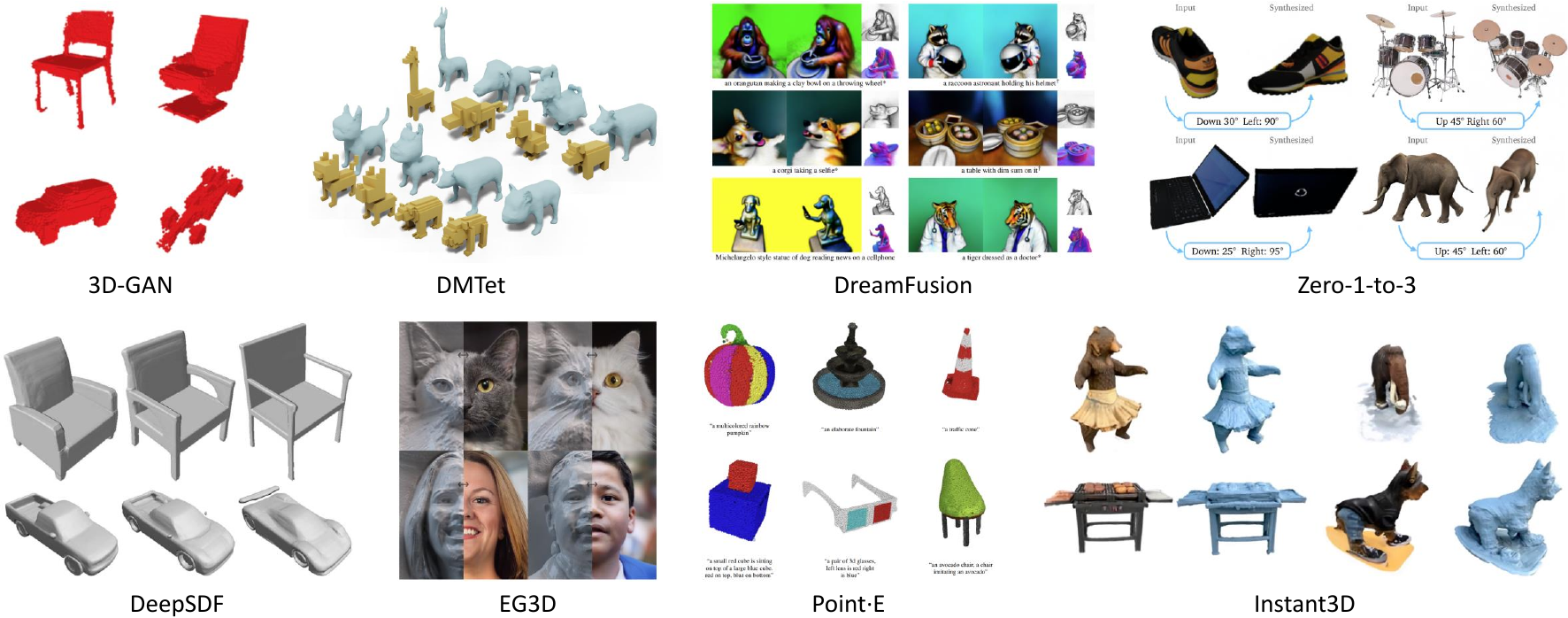}
\end{center}
\caption{In this survey, we investigate a large variety of 3D generation methods. Over the past decade, 3D generation has achieved remarkable progress and has recently garnered considerable attention due to the success of generative AI in images and videos. 3D generation results from 3D-GAN~\cite{wu2016learning}, DeepSDF~\cite{park2019deepsdf}, DMTet~\cite{shen2021deep}, EG3D~\cite{chan2022efficient}, DreamFusion~\cite{poole2022dreamfusion}, PointE~\cite{nichol2022point}, Zero-1-to-3~\cite{liu2023zero} and Instant3D~\cite{li2023instant3d}.}
\label{fig:teaser}
\end{figure*}

{\section{Introduction}\label{sec:intro}}

Automatically generating 3D models using algorithms has long been a significant task in computer vision and graphics. This task has garnered considerable interest due to its broad applications in video games, movies, virtual characters, and immersive experiences, which typically require a wealth of 3D assets. Recently, the success of neural representations, particularly Neural Radiance Fields (NeRFs) \cite{mildenhall2020nerf, barron2021mip, muller2022instant, kerbl20233d}, and generative models such as diffusion models~\cite{ho2020denoising,rombach2022ldm}, has led to remarkable advancements in 3D content generation. 

In the realm of 2D content generation, recent advancements in generative models have steadily enhanced the capacity for image generation and editing, leading to increasingly diverse and high-quality results. Pioneering research on generative adversarial networks (GANs) \cite{goodfellow2014generative, abdal2019image2stylegan}, variational autoencoders (VAEs) \cite{kusner2017grammar, pu2016variational, vae}, and autoregressive models \cite{gpt2, gpt3} has demonstrated impressive outcomes. Furthermore, the advent of generative artificial intelligence (AI) and diffusion models \cite{ho2020denoising, nichol2021improved, saharia2022photorealistic} signifies a paradigm shift in image manipulation techniques, such as Stable Diffusion \cite{rombach2022ldm}, Imagen \cite{saharia2022photorealistic}, Midjourney \cite{Midjourney}, or DALL-E 3 \cite{dalle3}. These generative AI models enable the creation and editing of photorealistic or stylized images, or even videos \cite{chen2024videocrafter2, ho2022video,singer2022make, ge2023preserve}, using minimal input like text prompts. As a result, they often generate imaginative content that transcends the boundaries of the real world, pushing the limits of creativity and artistic expression. Owing to their ``emergent'' capabilities, these models have redefined the limits of what is achievable in content generation, expanding the horizons of creativity and artistic expression.


The demand to extend 2D content generation into 3D space is becoming increasingly crucial for applications in generating 3D assets or creating immersive experiences, particularly with the rapid development of the metaverse. The transition from 2D to 3D content generation, however, is not merely a technological evolution. It is primarily a response to the demands of modern applications that necessitate a more intricate replication of the physical world, which 2D representations often fail to provide. This shift highlights the limitations of 2D content in applications that require a comprehensive understanding of spatial relationships and depth perception. 

As the significance of 3D content becomes increasingly evident, there has been a surge in research efforts dedicated to this domain. However, the transition from 2D to 3D content generation is not a straightforward extension of existing 2D methodologies. Instead, it involves tackling unique challenges and re-evaluating data representation, formulation, and underlying generative models to effectively address the complexities of 3D space. For instance, it is not obvious how to integrate the 3D scene representations into 2D generative models to handle higher dimensions, as required for 3D generation. Unlike images or videos which can be easily collected from the web, 3D assets are relatively scarce. Furthermore, evaluating the quality of generated 3D models presents additional challenges, as it is necessary to develop better formulations for objective functions, particularly when considering multi-view consistency in 3D space. These complexities demand innovative approaches and novel solutions to bridge the gap between 2D and 3D content generation.

While not as prominently featured as its 2D counterpart, 3D content generation has been steadily progressing with a series of notable achievements. The representative examples shown in Fig. \ref{fig:teaser} demonstrate significant improvements in both quality and diversity, transitioning from early methods like 3D-GAN~\cite{wu2016learning} to recent approaches like Instant3D~\cite{li2023instant3d}. Therefore, This survey paper seeks to systematically explore the rapid advancements and multifaceted developments in 3D content generation. We present a structured overview and comprehensive roadmap of the many recent works focusing on 3D representations, 3D generation methods, datasets, and applications of 3D content generation, and to outline open challenges. 



\begin{figure*}[t]
\begin{center}
   \includegraphics[width=0.95\linewidth]{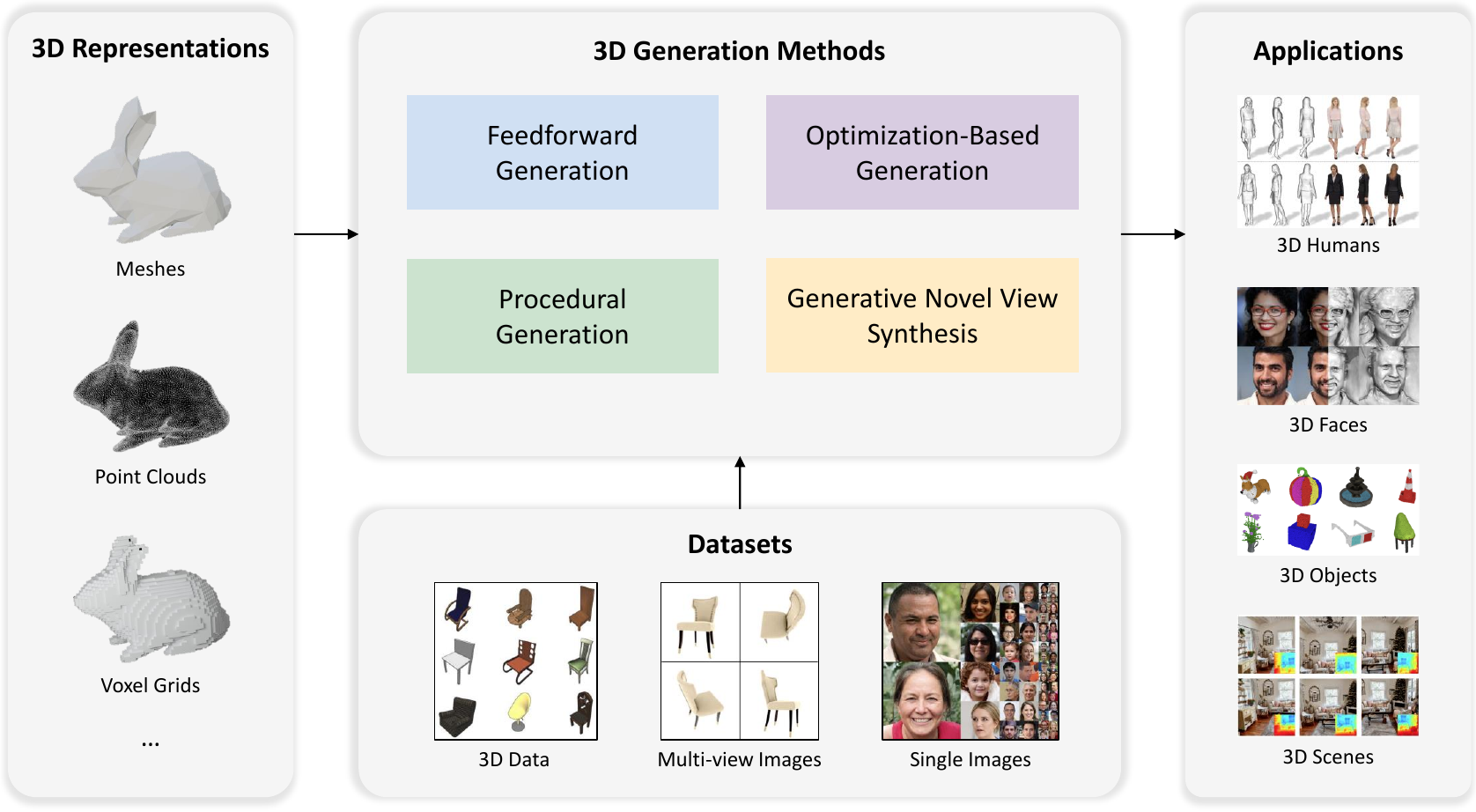}
\end{center}
\caption{Overview of this survey, including 3D representations, 3D generation methods, datasets and applications. Specifically, we introduce the 3D representations that serve as the backbone for 3D generation. Furthermore, we provide a comprehensive overview of the rapidly growing literature on generation methods, categorized by the type of algorithmic paradigms, including feedforward generation, optimization-based generation, procedural generation, and generative novel view synthesis. Finally, we provide a brief discussion on popular datasets and available applications.}
\label{fig:overview}
\end{figure*}

Fig. \ref{fig:overview} presents an overview of this survey. We first discuss the scope and related work of this survey in Sec.~\ref{sec:scope}. In the following sections, we examine the core methodologies that form the foundation of 3D content generation. Sec. \ref{sec:representation} introduces the primary scene representations and their corresponding rendering functions used in 3D content generation. Sec. \ref{sec:generation} explores a wide variety of 3D generation methods, which can be divided into four categories based on their algorithmic methodologies: feedforward generation, optimization-based generation, procedural generation, and generative novel view synthesis. An evolutionary tree of these methodologies is also depicted to illustrate their primary branch. As data accumulation plays a vital role in ensuring the success of deep learning models, we present related datasets employed for training 3D generation methods. In the end, we include a brief discussion on related applications, such as 3D human and face generation, outline open challenges, and conclude this survey. We hope this survey offers a systematic summary of 3D generation that could inspire subsequent work for interested readers.

In this work, we present a comprehensive survey on 3D generation, with two main contributions:
\begin{itemize}
    \item Given the recent surge in contributions based on generative models in the field of 3D vision, we provide a comprehensive and timely literature review of 3D content generation, aiming to offer readers a rapid understanding of the 3D generation framework and its underlying principles.  
    
    \item We propose a multi-perspective categorization of 3D generation methods, aiming to assist researchers working on 3D content generation in specific domains to quickly identify relevant works and facilitate a better understanding of the related techniques. 
\end{itemize}

\section{Scope of This Survey} \label{sec:scope}
In this survey, we concentrate on the techniques for the generation of 3D models and their related datasets and applications. Specifically, we first give a short introduction to the scene representation. Our focus then shifts to the integration of these representations and the generative models. Then, we provide a comprehensive overview of the prominent methodologies of generation methods. We also explore the related datasets and cutting-edge applications such as 3D human generation, 3D face generation, and 3D editing, all of which are enhanced by these techniques. 

This survey is dedicated to systematically summarizing and categorizing 3D generation methods, along with the related datasets and applications. The surveyed papers are mostly published in major computer vision and computer graphics conferences/journals as well as some preprints released on arXiv in 2023. While it's challenging to exhaust all methods related to 3D generation, we hope to include as many major branches of 3D generation as possible. We do not delve into detailed explanations for each branch, instead, we typically introduce some representative works within it to explain its paradigm. The details of each branch can be found in the related work section of these cited papers.

\noindent\textbf{Related Survey.} Neural reconstruction and rendering with scene representations are closely related to 3D generation. However, we consider these topics to be outside the purview of this report. For a comprehensive discussion on neural rendering, we direct readers to \cite{tewari2020state, tewari2022advances}, and for a broader examination of other neural representations, we recommend \cite{kato2020differentiable, xie2022neural}. Our primary focus is on exploring techniques that generate 3D models. Therefore, this review does not encompass research on generation methods for 2D images within the realm of visual computing. For further information on a specific generation method, readers can refer to \cite{doersch2016tutorial} (VAEs), \cite{gui2021review} (GANs), \cite{po2023state, croitoru2023diffusion} (Diffusion) and \cite{khan2022transformers} (Transformers) for a more detailed understanding. There are also some surveys related to 3D generation that have their own focuses such as 3D-aware image synthesis~\cite{xia2023survey}, 3D generative models~\cite{shi2022deep}, Text-to-3D~\cite{li2023generative} and deep learning for 3D point clouds~\cite{guo2020deep}. In this survey, we give a comprehensive analysis of different 3D generation methods.

\section{Neural Scene Representations}\label{sec:representation}

\begin{figure*}
    \centering
    \includegraphics[width=.95\linewidth]{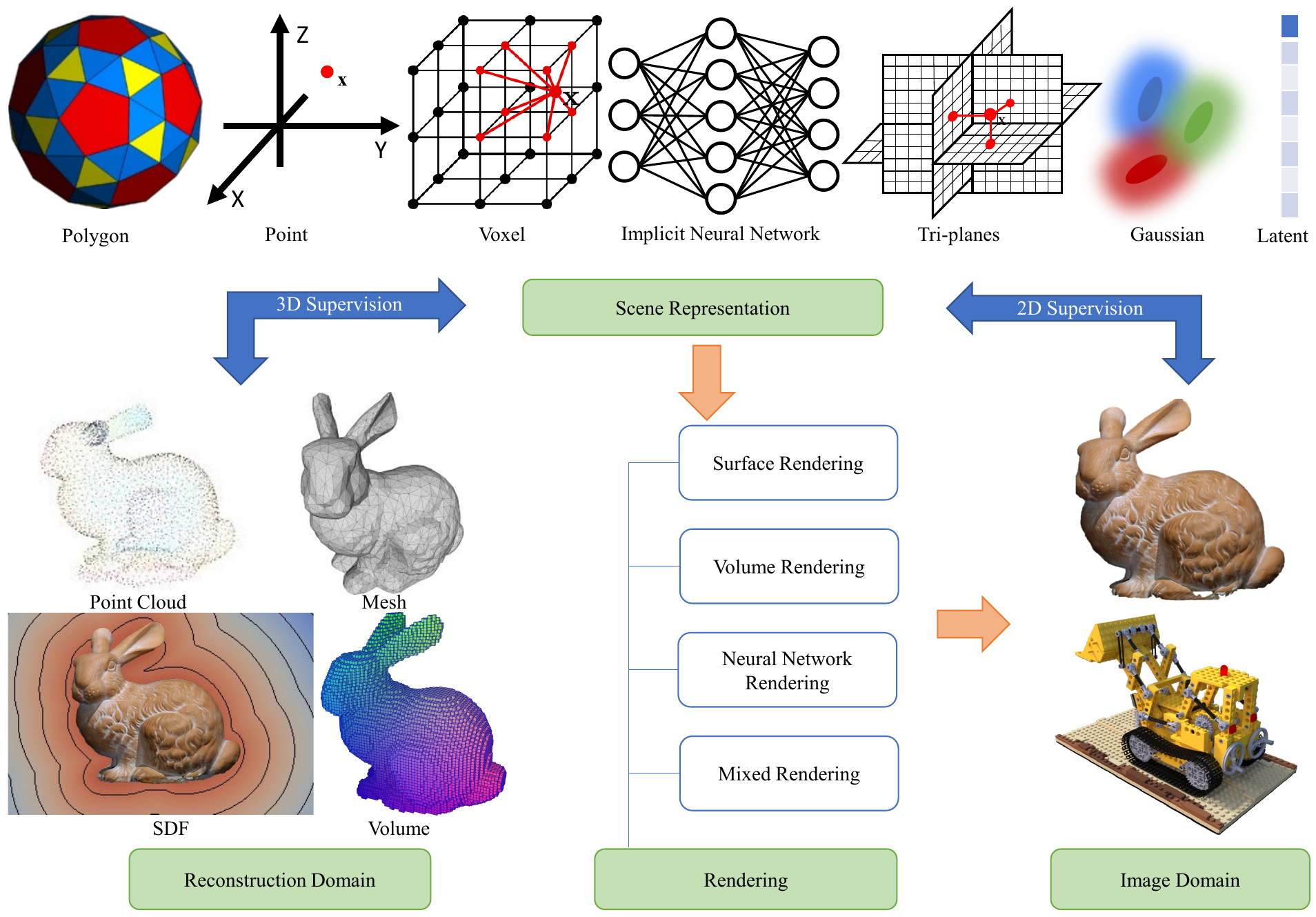}
    \caption{Neural scene representations used for 3D generation, including explicit, implicit, and hybrid representations. The 3D generation involves the use of scene representations and a differentiable rendering algorithm to create 3D models or render 2D images. On the flip side, these 3D models or 2D images can function as the reconstruction domain or image domain, overseeing the 3D generation of scene representations.}
    \label{fig:neural_scene}
\end{figure*}

In the domain of 3D AI-generated content, adopting a suitable representation of 3D models is essential. The generation process typically involves a scene representation and a differentiable rendering algorithm for creating 3D models and rendering 2D images. Conversely, the created 3D models or 2D images could be supervised in the reconstruction domain or image domain, as illustrated in Fig. \ref{fig:neural_scene}. Some methods directly supervise the 3D models of the scene representation, while others render the scene representation into images and supervise the resulting renderings. In the following, we broadly classify the scene representations into three groups: explicit scene representations (Section \ref{subsec:explicit_rep}), implicit representations (Section \ref{subsec:implcit_rep}), and hybrid representations (Section \ref{subsec:hybrid_rep}). Note that, the rendering methods (\textit{e.g.} ray casting, volume rendering, rasterization, \textit{etc}), which should be differentiable to optimize the scene representations from various inputs, are also introduced.


\subsection{Explicit Representations} \label{subsec:explicit_rep}




Explicit scene representations serve as a fundamental module in computer graphics and vision, as they offer a comprehensive means of describing 3D scenes. By depicting scenes as an assembly of basic primitives, including point-like primitives, triangle-based meshes, and advanced parametric surfaces, these representations can create detailed and accurate visualizations of various environments and objects. 

\subsubsection{Point Clouds}
A point cloud is a collection of elements in Euclidean space, representing discrete points with addition attributes (\textit{e.g.} colors and normals) in three-dimensional space. In addition to simple points, which can be considered infinitesimally small surface patches, oriented point clouds with a radius (surfels) can also be used \cite{pfister2000surfels}.
Surfels are used in computer graphics for rendering point clouds (called splitting), which are differentiable \cite{yifan2019differentiable, kerbl20233d} and allow researchers to define differentiable rendering pipelines to adjust point cloud positions and features, such as radius or color. 
Techniques like Neural Point-based Rendering \cite{aliev2020neural, dai2020neural}, SynSin \cite{wiles2020synsin}, Pulsar \cite{lassner2021pulsar, kopanas2021point} and ADOP \cite{ruckert2022adop}  leverage learnable features to store information about the surface appearance and shape, enabling more accurate and detailed rendering results. Several other methods, such as FVS \cite{riegler2020free}, SVS \cite{riegler2021stable}, and FWD-Transformer \cite{cao2022fwd}, also employ learnable features to improve the rendering quality. These methods typically embed features into point clouds and warp them to target views to decode color values, allowing for more accurate and detailed reconstructions of the scene.

By incorporating point cloud-based differentiable renderers into the 3D generation process, researchers can leverage the benefits of point clouds while maintaining compatibility with gradient-based optimization techniques. This process can be generally categorized into two different ways: point splitting which blends the discrete samples with some local deterministic blurring kernels \cite{zwicker2002ewa, lin2018learning, insafutdinov2018unsupervised, roveri2018network}, and conventional point renderer \cite{aliev2020neural, dai2020neural,kopanas2021point,rakhimov2022npbg++}. These methods facilitate the generation and manipulation of 3D point cloud models while maintaining differentiability, which is essential for training and optimizing neural networks in 3D generation tasks.

\subsubsection{Meshes}
By connecting multiple vertices with edges, more complex geometric structures (\textit{e.g.} wireframes and meshes) can be formed \cite{botsch2010polygon}. These structures can then be further refined by using polygons, typically triangles or quadrilaterals, to create realistic representations of objects \cite{shirman1987local}. Meshes provide a versatile and efficient means of representing intricate shapes and structures, as they can be easily manipulated and rendered by computer algorithms. 
The majority of graphic editing toolchains utilize triangle meshes. This type of representation is indispensable for any digital content creation (DCC) pipeline, given its wide acceptance and compatibility. To align seamlessly with these pipelines, neural networks can be strategically trained to predict discrete vertex locations \cite{burov2021dynamic, thies2019deferred}. This ability allows for the direct importation of these locations into any DCC pipeline, facilitating a smooth and efficient workflow. In contrast to predicting discrete textures, continuous texture methods optimized via neural networks are proposed, such as texture fields \cite{oechsle2019texture} and NeRF-Tex \cite{baatz2022nerf}. In this way, it could provide a more refined and detailed texture, enhancing the overall quality and realism of the generated 2D models.

Integrating mesh representation into 3D generation requires the use of mesh-based differentiable rendering methods, which enable meshes to be rasterized in a manner that is compatible with gradient-based optimization. Several such techniques have been proposed, including OpenDR \cite{loper2014opendr}, neural mesh renderer \cite{kato2018neural}, Paparazzi \cite{liu2018paparazzi}, and Soft Rasterizer \cite{liu2019soft}. Additionally, general-purpose physically based renderers like Mitsuba 2 \cite{nimier2019mitsuba} and Taichi \cite{hu2019taichi} support mesh-based differentiable rendering through automatic differentiation.

\subsubsection{Multi-layer Representations}
The use of multiple semi-transparent colored layers for representing scenes has been a popular and successful scheme in real-time novel view synthesis \cite{zhou2018stereo}. Using Layered Depth Image (LDI) representation \cite{shade1998layered} is a notable example, extending traditional depth maps by incorporating multiple layers of depth maps, each with associated color values. Several methods \cite{penner2017soft, choi2019extreme, shih20203d} have drawn inspiration from the LDI representation and employed deep learning advancements to create networks capable of predicting LDIs. In addition to LDIs, Stereomagnification \cite{zhou2018stereo} initially introduced the multiple image (MPI) representation. It describes scenes using multiple front-parallel semi-transparent layers, including colors and opacity, at fixed depth ranges through plane sweep volumes. With the help of volume rendering and homography projection, the novel view could be synthesized in real-time. Building on Stereomagnification \cite{zhou2018stereo}, various methods \cite{flynn2019deepview, mildenhall2019local, srinivasan2019pushing} have adopted the MPI representation to enhance rendering quality. The multi-layer representation has been further expanded to accommodate wider fields of view in \cite{broxton2020immersive, attal2020matryodshka, lin2020deep} by substituting planes with spheres. As research in this domain continues to evolve, we can expect further advancements in these methods, leading to more efficient and effective 3D generation techniques for real-time rendering.

\subsection{Implicit Representations} \label{subsec:implcit_rep}
Implicit representations have become the scene representation of choice for problems in view synthesis or shape reconstruction, as well as many other applications across computer graphics and vision. Unlike explicit scene representations that usually focus on object surfaces, implicit representations could define the entire volume of a 3D object, and use volume rendering for image synthesis. These representations utilize mathematical functions, such as radiance fields \cite{mildenhall2020nerf} or signed distance fields \cite{park2019deepsdf, chen2019learning}, to describe the properties of a 3D space. 

\subsubsection{Neural Radiance Fields}
Neural Radiance Fields (NeRFs) \cite{mildenhall2020nerf} have gained prominence as a favored scene representation method for a wide range of applications. Fundamentally, NeRFs introduce a novel representation of 3D scenes or geometries. Rather than utilizing point clouds and meshes, NeRFs depict the scene as a continuous volume. This approach involves obtaining volumetric parameters, such as view-dependent radiance and volume density, by querying an implicit neural network. This innovative representation offers a more fluid and adaptable way to capture the intricacies of 3D scenes, paving the way for enhanced rendering and modeling techniques.


Specifically, NeRF \cite{mildenhall2020nerf} represents the scene with a continuous volumetric radiance field, which utilizes MLPs to map the position  $\mathbf{x}$ and view direction $\mathbf{r}$ to a density $\sigma$ and color $\mathbf{c}$. To render a pixel's color, NeRF casts a single ray $\mathbf{r}(t)=\mathbf{o}+t\mathbf{d}$ and evaluates a series of points $\{t_i\}$ along the ray. The evaluated $\{(\sigma_i, \mathbf{c}_i)\}$ at the sampled points are accumulated into the color ${C}(\mathbf{r})$ of the pixel via volume rendering \cite{max1995optical}:
\begin{equation}\small
     {C(r)}\!=\!\sum_{i} T_i \alpha_i \mathbf{c}_i,\quad 
     \text{where}\  T_i = \exp  \left({-\sum_{k=0}^{i-1}\sigma_{k}\delta_{k}}\right),
    \label{eq:vol_ren} 
\end{equation}
and $\alpha_i=1-\exp(-\sigma_i\delta_{i})$ indicates the opacity of the sampled point. Accumulated transmittance $T_i$ quantifies the probability of the ray traveling from $t_0$ to $t_i$ without encountering other particles, and $\delta_{i} = t_i-t_{i-1}$ denotes the distance between adjacent samples. 

NeRFs \cite{mildenhall2020nerf, niemeyer2021giraffe, barron2021mip, barron2022mip, verbin2022ref, li2023dynibar} have seen widespread success in problems such as edition \cite{martin2021nerf, zhi2021place, chen2022hallucinated, yuan2022nerf}, joint optimization of cameras \cite{lin2021barf, wang2021nerf, chen2023local, truong2023sparf}, inverse rendering \cite{zhang2021physg, srinivasan2021nerv, boss2021nerd, zhang2021nerfactor, zhuang2023neai, liang2023gs}, generalization \cite{yu2021pixelnerf, wang2021ibrnet, chen2021mvsnerf,li2021mine, johari2022geonerf, huang2023local}, acceleration \cite{reiser2021kilonerf, garbin2021fastnerf, zhu2023pyramid}, and free-viewpoint video \cite{du2021neural, li2022neural, pumarola2021d}. Apart from the above applications, NeRF-based representation can also be used for digit avatar generation, such as face and body reenactment \cite{peng2021animatable, guo2021ad, liu2021neural, weng2022humannerf, hong2022headnerf}.
NeRFs have been extend to various fields such as robotics \cite{kerr2022evo, zhou2023nerf, adamkiewicz2022vision}, tomography \cite{ruckert2022neat, zhu2022dnf}, image processing \cite{huang2022hdr, ma2022deblur, huang2023inverting}, and astronomy \cite{levis2022gravitationally}. 

\subsubsection{Neural Implicit Surfaces}

Within the scope of shape reconstruction, a neural network processes a 3D coordinate as input and generates a scalar value, which usually signifies the signed distance to the surface. This method is particularly effective in filling in missing information and generating smooth, continuous surfaces. The implicit surface representation defines the scene's surface as a learnable function $f$ that specifies the signed distance $f(\mathbf{x})$ from each point to the surface. The fundamental surface can then be extracted from the zero-level set,
$S=\{\mathbf{x}\in\mathbb{R}^3|f(\mathbf{x})=0\}$, providing a flexible and efficient way to reconstruct complex 3D shapes.
Implicit surface representations offer numerous advantages, as they eliminate the need to define mesh templates. As a result, they can represent objects with unknown or changing topology in dynamic scenarios. Specifically, implicit surface representations recover signed distance fields for shape modeling using MLPs with coordinate inputs. These initial proposals sparked widespread enthusiasm and led to various improvements focusing on different aspects, such as enhancing training schemes \cite{duan2020curriculum, yifan2020neural, zhou2022learning}, leveraging global-local context \cite{xu2019disn, erler2020points2surf, zhu2022nice}, adopting specific parameterizations \cite{genova2019learning, chen2020bsp, yifan2021geometry, ben2022digs}, and employing spatial partitions \cite{genova2020local, tretschk2020patchnets, takikawa2021neural, wang2023lp}.

NeuS \cite{wang2021neus} and VolSDF \cite{yariv2021volume} extend the basic NeRF formulation by integrating an SDF into volume rendering, which defines a function to map the signed distance to density $\sigma$. It attains a locally maximal value at surface intersection points. Specifically, accumulated transmittance $T(t)$ along the ray $\mathbf{r}(t)=\mathbf{o}+t\mathbf{d}$ is formulated as a sigmoid function: $T(t)=\Phi(f(t))=(1+e^{sf(t)})^{-1}$, where $s$ and $f(t)$ refer to a learnable parameter and the signed distance function of points at $\mathbf{r}(t)$, respectively. Discrete opacity values $\alpha_i$ can then be derived as:
\begin{equation}
\alpha_i=\max\left(
    \frac{
        \Phi_s\left(f(t_i)\right) - \Phi_s\left(f(t_{i+1})\right)
    }{
        \Phi_s\left(f(t_i)\right)
    }, 0
\right).
\label{eq:neus_opacity}
\end{equation}
NeuS employs volume rendering to recover the underlying SDF based on Eqs. \eqref{eq:vol_ren} and \eqref{eq:neus_opacity}. The SDF is optimized by minimizing the photometric loss between the rendering results and ground-truth images.

Building upon NeuS and VolSDF, NeuralWarp \cite{darmon2022improving}, Geo-NeuS \cite{fu2022geo}, MonoSDF \cite{yu2022monosdf} leverage prior geometry information from MVS methods. IRON \cite{zhang2022iron}, MII \cite{zhang2022modeling}, and WildLight \cite{cheng2023wildlight} apply high-fidelity shape reconstruction via SDF for inverse rendering. HF-NeuS \cite{wang2022hf} and PET-Neus \cite{wang2023petneus} integrate additional displacement networks to fit the high-frequency details. LoD-NeuS \cite{zhuang2023anti} adaptively encodes Level of Detail (LoD) features for shape reconstruction. 

\begin{table*}[ht]
    \centering
    \footnotesize
    \setlength{\tabcolsep}{4.4pt} 
    \caption{Some examples of 3D generation methods. We first divide the methods according to the generative models and their corresponding representations in generation space. The representations in the reconstruction space determine how the 3D objects are formatted and rendered. We also list the main supervision and conditions of these methods. For the 2D supervision, a rendering technique is utilized to generate the images.}
    \begin{tabular}{@{}lccccccc@{}}
    \toprule
    Method & Generative Model & Generation Space & \thead{Reconstruction Space} & Rendering & Supervision & Condition \\
    \midrule

    PointFlow~\cite{pointflow} & Normalizing Flow & Latent Code & Point Cloud & - & 3D & Uncon \\
    
    3dAAE~\cite{zamorski2020adversarial} & VAE & Latent Code & Point Cloud & - & 3D & Uncon \\
    SDM-NET~\cite{gao2019sdm} & VAE & Latent Code & Mesh & - & 3D & Uncon \\

    \midrule
    
    AutoSDF~\cite{mittal2022autosdf} & Autoregressive  & Voxel & SDF & - & 3D & Uncon. \\
    PolyGen~\cite{nash2020polygen} & Autoregressive  & Polygon & Mesh & - & 3D & Uncon./Label/Image \\
    PointGrow~\cite{sun2020pointgrow} & Autoregressive  & Point & Point Cloud & - & 3D & Uncon./Label/Image \\

    \midrule
    
    EG3D~\cite{chan2022efficient} &  GAN  & Latent Code & Tri-plane & Mixed Rendering & 2D & Uncon. \\
    GIRAFFE~\cite{niemeyer2021giraffe} & GAN & Latent Code & NeRF & Mixed Rendering & 2D & Uncon. \\
    BlockGAN~\cite{nguyen2020blockgan} & GAN & Latent Code & Voxel Grid & Network Rendering & 2D & Uncon. \\
    gDNA~\cite{chen2022gdna} & GAN & Latent Code & Occupancy Field & Surface Rendering & 2D\&3D & Uncon. \\
    SurfGen~\cite{luo2021surfgen} & GAN & Latent Code & SDF & - & 3D & Uncon. \\
    tree-GAN~\cite{shu20193d} & GAN & Latent Code & Point Cloud & - & 3D & Uncon. \\

    \midrule
    HoloDiffusion~\cite{karnewar2023holodiffusion} & Diffusion & Voxel & NeRF & Volume Rendering & 2D & Image\\
    SSDNeRF~\cite{chen2023single} & Diffusion & Tri-plane & NeRF & Volume Rendering & 2D & Uncon./Image\\
    3DShape2VecSet~\cite{zhang20233dshape2vecset} & Diffusion & Latent Set & SDF & - & 3D & Uncon./Text/Image \\
    Point-E~\cite{nichol2022point} & Diffusion & Point & Point Cloud & - & 3D & Text\\
    3DGen~\cite{gupta20233dgen} & Diffusion & Tri-plane & Mesh & - & 3D & Text/Image\\
    
    DreamFusion~\cite{poole2022dreamfusion} & Diffusion & -  & NeRF & Volume Rendering & SDS & Text \\
    Make-It-3D~\cite{tang2023make} & Diffusion & -  & Point Cloud & Network Rendering & SDS & Image \\
    Zero-1-to-3~\cite{liu2023zero} & Diffusion & Pixel  & - & - & 2D & Image \\
    MVDream~\cite{shi2023mvdream} & Diffusion & Pixel  & - & - & 2D & Image \\
    DMV3D~\cite{xu2023dmv3d} & Diffusion & Pixel & Tri-plane & Volume Rendering & 2D & Text/Image\\

    \bottomrule
    \end{tabular}
    \label{tab:example_methods}
\end{table*}

\subsection{Hybrid Representations}\label{subsec:hybrid_rep}
Implicit representations have indeed demonstrated impressive results in various applications as mentioned above. However, most of the current implicit methods rely on regression to NeRF or SDF values, which may limit their ability to benefit from explicit supervision on the target views or surfaces. Explicit representation could impose useful constraints during training and improve the user experience. To capitalize on the complementary benefits of both representations, researchers have begun exploring hybrid representations. These involve scene representations (either explicit or implicit) that embed features utilizing rendering algorithms for view synthesis.   
\subsubsection{Voxel Grids}

Early work \cite{wu20153d, choy20163d, maturana2015voxnet} depicted 3D shapes using voxels, which store coarse occupancy (inside/outside) values on a regular grid. This approach enabled powerful convolutional neural networks to operate natively and produce impressive results in 3D reconstruction and synthesis \cite{dai2018scancomplete, wu2016learning, brock2016generative}. These methods usually use explicit voxel grids as the 3D representation. Recently, to address the slow training and rendering speeds of implicit representations, the 3D voxel-based embedding methods \cite{liu2020neural, fridovich2022plenoxels, schwarz2022voxgraf, sun2022direct} have been proposed. These methods encode the spatial information of the scene and decode the features more efficiently. Moreover, Instant-NGP \cite{muller2022instant} introduces the multi-level voxel grids encoded implicitly via the hash function for each level. It facilitates rapid optimization and rendering while maintaining a compact model. These advancements in 3D shape representation and processing techniques have significantly enhanced the efficiency and effectiveness of 3D generation applications.

\subsubsection{Tri-plane}
Tri-plane representation is an alternative approach to using voxel grids for embedding features in 3D shape representation and neural rendering. The main idea behind this method is to decompose a 3D volume into three orthogonal planes (e.g., XY, XZ, and YZ planes) and represent the features of the 3D shape on these planes. Specifically, TensoRF \cite{chen2022tensorf} achieves similar model compression and acceleration by replacing each voxel grid with a tensor decomposition into planes and vectors. Tri-planes are efficient and capable of scaling with the surface area rather than volume and naturally integrate with expressive, fine-tuned 2D generative architectures. In the generative setting, EG3D \cite{chan2022efficient} proposes a spatial decomposition into three planes whose values are added together to represent a 3D volume. NFD \cite{shue20233d} introduces diffusion on 3D scenes, utilizing 2D diffusion model backbones and having built-in tri-plane representation.

\subsubsection{Hybrid Surface Representation}

DMTet, a recent development cited in \cite{shen2021deep}, is a hybrid three-dimensional surface representation that combines both explicit and implicit forms to create a versatile and efficient model. It segments the 3D space into dense tetrahedra, thereby forming an explicit partition. By integrating explicit and implicit representations, DMTet can be optimized more efficiently and transformed seamlessly into explicit structures like mesh representations. During the generation process, DMTet can be differentiably converted into a mesh, which enables swift high-resolution multi-view rendering. This innovative approach offers significant improvements in terms of efficiency and versatility in 3D modeling and rendering.



\section{Generation Methods} \label{sec:generation}

In the past few years, the rapid development of generative models in 2D image synthesis, such as generative adversarial networks (GANs) \cite{goodfellow2014generative, abdal2019image2stylegan}, variational autoencoders (VAEs) \cite{kusner2017grammar, pu2016variational, vae}, autoregressive models\cite{gpt2, gpt3}, diffusion models \cite{ho2020denoising, nichol2021improved, saharia2022photorealistic}, \textit{etc}., has led to their extension and combination with these scene representations for 3D generation. Tab. \ref{tab:example_methods} shows well-known examples of 3D generation using generative models and scene representations. These methods may use different scene representations in the generation space, where the representation is generated by the generative models, and the reconstruction space, where the output is represented. For example, AutoSDF~\cite{mittal2022autosdf} uses a transformer-based autoregressive model to learn a feature voxel grid and decode this representation to SDF for reconstruction. EG3D \cite{chan2022efficient} employs GANs to generate samples in latent space and introduces a tri-plane representation for rendering the output. SSDNeRF~\cite{chen2023single} uses the diffusion model to generate tri-plane features and decode them to NeRF for rendering. By leveraging the advantages of neural scene representations and generative models, these approaches have demonstrated remarkable potential in generating realistic and intricate 3D models while maintaining view consistency. 

\begin{figure*}[p]
\begin{center}
   \includegraphics[width=1.0\linewidth]{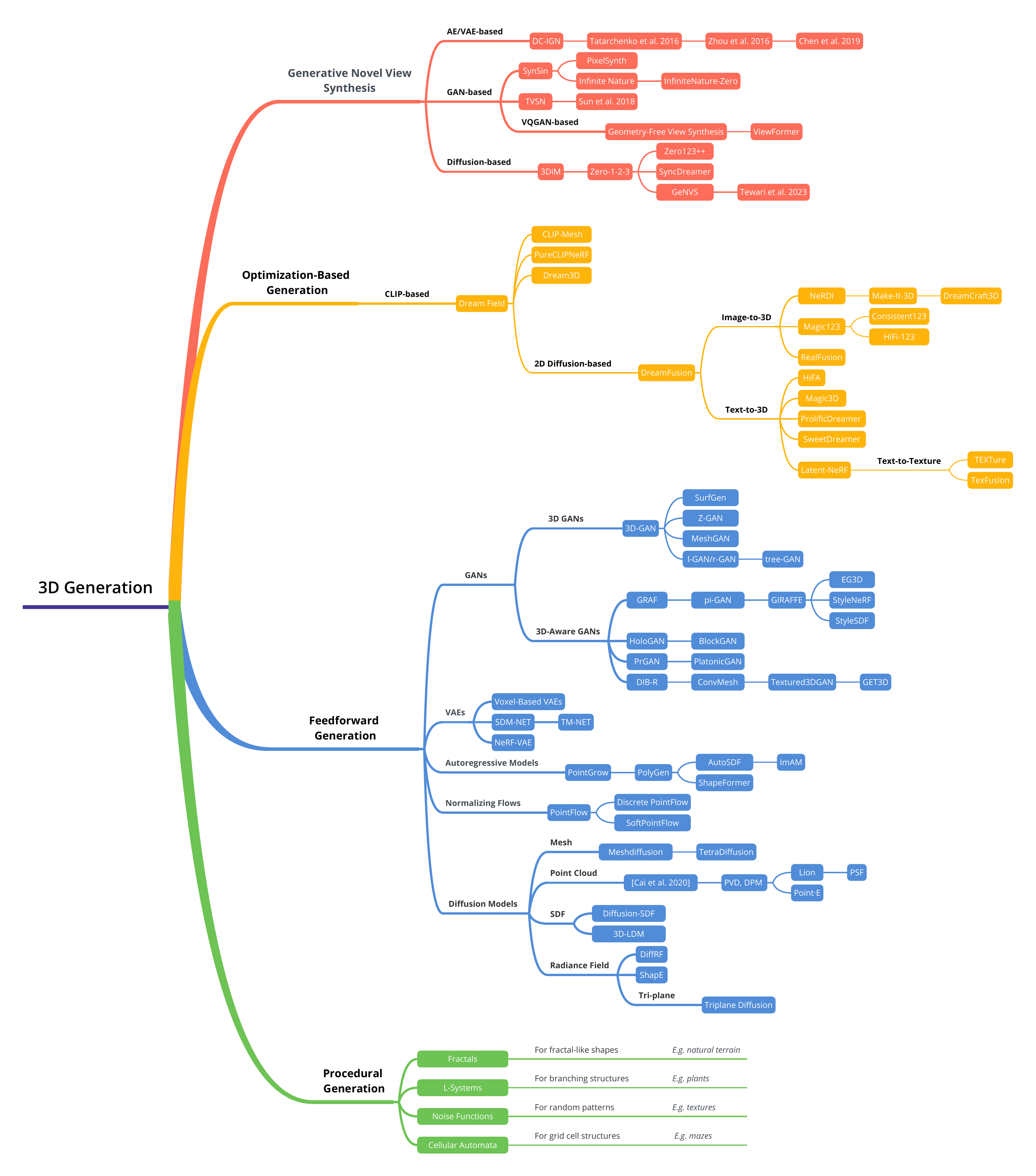}
\end{center}
\caption{The evolutionary tree of 3D generation illustrates the primary branch of generation methods and their developments in recent years. Specifically, we provide a comprehensive overview of the rapidly growing literature on generation methods, categorized by the type of algorithmic paradigms, including feedforward generation, optimization-based generation, procedural generation, and generative novel view synthesis.}
\label{fig:evolutionarytree}
\end{figure*}

In this section, we explore a large variety of 3D generation methods which are organized into four categories based on their algorithmic paradigms: Feedforward Generation (Sec.~\ref{sec:ff_gen}), generating results in a forward pass; Optimization-Based Generation (Sec.~\ref{sec:opt_gen}), necessitating a test-time optimization for each generation; Procedural Generation (Sec.~\ref{sec:pro_gen}), creating 3D models from sets of rules; and Generative Novel View Synthesis (Sec.~\ref{sec:gnvs}), synthesizing multi-view images rather than an explicit 3D representation for 3D generation. An evolutionary tree of 3D generation methods is depicted in Fig.~\ref{fig:evolutionarytree}, which illustrates the primary branch of generation techniques, along with associated work and subsequent developments. A comprehensive analysis will be discussed in the subsequent subsection. 

\subsection{Feedforward Generation}
\label{sec:ff_gen}

\begin{figure*}
    \centering
    \includegraphics[width=1.0\linewidth]{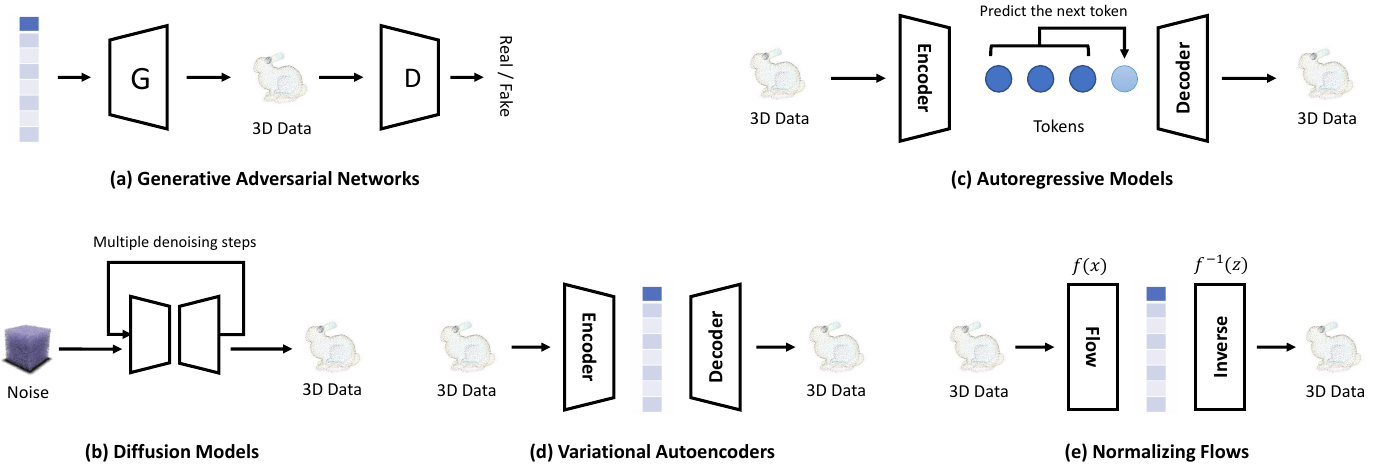}
    \caption{Exemplary feedforward 3D generation models. We showcase several representative pipelines of feedforward 3D generation models, including (a) generative adversarial networks, (b) diffusion models, (c) autoregressive models, (d) variational autoencoders and (e) normalizing flows.}
    \label{fig:models}
\end{figure*}

A primary technical approach for generation methods is feedforward generation, which can directly produce 3D representations using generative models. In this section, we explore these methods based on their generative models as shown in Fig.~\ref{fig:models}, which include generative adversarial networks (GANs), diffusion Models, autoregressive models, variational autoencoders (VAEs) and normalizing flows.

\subsubsection{Generative Adversarial Networks}

Generative Adversarial Networks (GANs)~\cite{goodfellow2014generative} have demonstrated remarkable outcomes in image synthesis tasks, consisting of a generator $G(\cdot)$ and a discriminator $D(\cdot)$. The generator network $G$ produces synthetic data by accepting latent code as input, while the discriminator network $D$ differentiates between generated data from $G$ and real data. Throughout the training optimization process, the generator $G$ and discriminator $D$ are jointly optimized, guiding the generator to create synthetic data as realistic as real data.

Building on the impressive results achieved by GANs in 2D image synthesis, researchers have begun to explore the application of these models to 3D generation tasks. The core idea is to marry GANs with various 3D representations, such as point clouds (l-GAN/r-GAN~\cite{achlioptas2018learning}, tree-GAN~\cite{shu20193d}), voxel grids (3D-GAN~\cite{wu2016learning}, Z-GAN~\cite{knyaz2018image}), meshes (MeshGAN~\cite{cheng2019meshgan}), or SDF (SurfGen~\cite{luo2021surfgen}, SDF-StyleGAN~\cite{zheng2022sdf}). In this context, the 3D generation process can be viewed as a series of adversarial steps, where the generator learns to create realistic 3D data from input latent codes, and the discriminator differentiates between generated data and real data. By iteratively optimizing the generator and discriminator networks, GANs learn to generate 3D data that closely resembles the realism of actual data. 

For 3D object generation, prior GAN methodologies, such as l-GAN~\cite{achlioptas2018learning}, 3D-GAN~\cite{wu2016learning}, and Multi-chart Generation~\cite{ben2018multi}, directly utilize explicit 3D object representation of real data to instruct generator networks. Their discriminators employ 3D representation as supervision, directing the generator to produce synthetic data that closely resembles the realism of actual data. During training, specialized generators generate corresponding supervisory 3D representations, such as point clouds, voxel grids, and meshes. Some studies, like SurfGen~\cite{luo2021surfgen}, have progressed further to generate intermediate implicit representations and then convert them to corresponding 3D representations instead of directly generating explicit ones, achieving superior performance. In particular, the generator of l-GAN~\cite{achlioptas2018learning}, 3D-GAN~\cite{wu2016learning}, and Multi-chart Generation~\cite{ben2018multi} generate the position of point cloud, voxel grid, and mesh directly, respectively, taking latent code as input. SurfGen~\cite{luo2021surfgen} generates implicit representation and then extracts explicit 3D representation. 

In addition to GANs that directly generate various 3D representations, researchers have suggested incorporating 2D supervision through differentiable rendering to guide 3D generation, which is commonly referred to as 3D-Aware GAN. Given the abundance of 2D images, GANs can better understand the implicit relationship between 2D and 3D data than relying solely on 3D supervision. In this approach, the generator of GANs generates rendered 2D images from implicit or explicit 3D representation. Then the discriminators distinguish between rendered 2D images and real 2D images to guide the training of the generator. 

Specifically, HoloGAN~\cite{nguyen2019hologan} first learns a 3D representation of 3D features, which is then projected to 2D features by the camera pose. These 2D feature maps are then rendered to generate the final images. BlockGAN~\cite{nguyen2020blockgan} extends it to generate 3D features of both background and foreground objects and combine them into 3D features for the whole scene. In addition, PrGAN~\cite{gadelha20173d} and PlatonicGAN~\cite{henzler2019escaping} employ an explicit voxel grid structure to represent 3D shapes and use a render layer to create images. Other methods like DIB-R~\cite{chen2019learning}, ConvMesh~\cite{pavllo2020convolutional}, Textured3DGAN~\cite{pavllo2021learning} and GET3D~\cite{gao2022get3d} propose GAN frameworks for generating triangle meshes and textures using only 2D supervision.

Building upon representations such as NeRFs, GRAF~\cite{schwarz2020graf} proposes generative radiance fields utilizing adversarial frameworks and achieves controllable image synthesis at high resolutions. pi-GAN~\cite{chan2021pi} introduces SIREN-based implicit GANs with FiLM conditioning to further improve image quality and view consistency. GIRAFFE~\cite{niemeyer2021giraffe} represents scenes as compositional generative neural feature fields to model multi-object scenes. Furthermore, EG3D~\cite{chan2022efficient} first proposes a hybrid explicit–implicit tri-plane representation that is both efficient and expressive and has been widely adopted in many following works.

\subsubsection{Diffusion Models}

Diffusion models~\cite{ho2020denoising,rombach2022ldm} are a class of generative models that learn to generate data samples by simulating a diffusion process. The key idea behind diffusion models is to transform the original data distribution into a simpler distribution, such as Gaussian, through a series of noise-driven steps called the forward process. The model then learns to reverse this process, known as the backward process, to generate new samples that resemble the original data distribution. The forward process can be thought of as gradually adding noise to the original data until it reaches the target distribution. The backward process, on the other hand, involves iteratively denoising the samples from the distribution to generate the final output. By learning this denoising process, diffusion models can effectively capture the underlying structure and patterns of the data, allowing them to generate high-quality and diverse samples. 

Building on the impressive results achieved by diffusion models in generating 2D images, researchers have begun to explore the applications of these models to 3D generation tasks. The core idea is to marry denoising diffusion models with various 3D representations. In this context, the 3D generation process can be viewed as a series of denoising steps, reversing the diffusion process from input 3D data to Gaussian noise. The diffusion models learn to generate 3D data from this noisy distribution through denoising. 


Specifically, Cai~\etal~\cite{cai2020learning} build upon a denoising score-matching framework to learn distributions for point cloud generation. PVD~\cite{zhou20213d} combines the benefits of both point-based and voxel-based representations for 3D generation. The model learns a diffusion process that transforms point clouds into voxel grids and vice versa, effectively capturing the underlying structure and patterns of the 3D data. Similarly, DPM~\cite{luo2021diffusion} focuses on learning a denoising process for point cloud data by iterative denoising the noisy point cloud samples. Following the advancements made by PVD~\cite{zhou20213d} and DPM~\cite{luo2021diffusion}, LION~\cite{zeng2022lion} builds upon the idea of denoising point clouds and introduces the concept of denoising in the latent space of point clouds, which is analogous to the shift in 2D image generation from denoising pixels to denoising latent space representations. To generate point clouds from text prompts, Point·E~\cite{nichol2022point} initially employs the GLIDE model~\cite{nichol2021glide} to generate text-conditional synthetic views, followed by the production of a point cloud using a diffusion model conditioned on the generated image. By training the model on a large-scale 3D dataset, it achieves remarkable generalization capabilities.

In addition to point clouds, MeshDiffusion~\cite{liu2023meshdiffusion}, Tetrahedral Diffusion Models~\cite{kalischek2022tetrahedral}, and SLIDE~\cite{lyu2023controllable} explore the application of diffusion models to mesh generation. MeshDiffusion~\cite{liu2023meshdiffusion} adopts the DMTet representation~\cite{shen2021deep} for meshes and optimizes the model by treating the optimization of signed distance functions as a denoising process. Tetrahedral Diffusion Models~\cite{kalischek2022tetrahedral} extends diffusion models to tetrahedral meshes, learning displacement vectors and signed distance values on the tetrahedral grid through denoising. SLIDE~\cite{lyu2023controllable} explores diffusion models on sparse latent points for mesh generation.

Apart from applying diffusion operations on explicit 3D representations, some works focus on performing the diffusion process on implicit representations. SSDNeRF~\cite{chen2023single}, DiffRF~\cite{muller2023diffrf} and Shap·E~\cite{jun2023shap} operate on 3D radiance fields, while SDF-Diffusion~\cite{shim2023diffusion}, LAS-Diffusion~\cite{zheng2023locally}, Neural Wavelet-domain Diffusion~\cite{hui2022neural}, One-2-3-45++~\cite{liu2023one}, SDFusion~\cite{cheng2023sdfusion} and 3D-LDM~\cite{nam20223d} focus on signed distance fields representations. Specifically, Diffusion-SDF~\cite{li2023diffusion} utilizes a voxel-shaped SDF representation to generate high-quality and continuous 3D shapes. 3D-LDM~\cite{nam20223d} creates neural implicit representations of SDFs by initially using a diffusion model to generate the latent space of an auto-decoder. Subsequently, the latent space is decoded into SDFs to acquire 3D shapes. Moreover, Rodin~\cite{wang2023rodin} and Shue~\etal~\cite{shue20233d} adopt tri-plane as the representation and optimize the tri-plane features using diffusion methods. Shue~\etal~\cite{shue20233d} generates 3D shapes using occupancy networks, while Rodin~\cite{wang2023rodin} obtains 3D shapes through volumetric rendering.



These approaches showcase the versatility of diffusion models in managing various 3D representations, including both explicit and implicit forms. By tailoring the denoising process to different representation types, diffusion models can effectively capture the underlying structure and patterns of 3D data, leading to improved generation quality and diversity. As research in this area continues to advance, it is expected that diffusion models will play a crucial role in pushing the boundaries of 3D shape generation across a wide range of applications. 

\subsubsection{Autoregressive Models}
A 3D object can be represented as a joint probability of the occurrences of multiple 3D elements:
\begin{equation}
    p(x_0, x_1, ..., x_n),
\end{equation}
where $x_i$ is the $i$-th element which can be the coordinate of a point or a voxel. A joint probability with a large number of random variables is usually hard to learn and estimate. However, one can factorize it into a product of conditional probabilities:
\begin{equation}
    p(x_0, x_1, ..., x_n) = p(x_0) \prod_{i=1}^{n} p(x_i | x_{<i}),
\end{equation}
which enables learning conditional probabilities and estimating the joint probability via sampling. Autoregressive models for data generation are a type of models that specify the current output depending on their previous outputs. Assuming that the elements $x_0$, $x_1$, ..., $x_n$ form an ordered sequence, a model can be trained by providing it with previous inputs $x_0$, ... $x_{i-1}$ and supervising it to fit the probability of the outcome $x_i$:
\begin{equation}
p(x_i | x_{<i}) = f(x_0, ..., x_{i-1}),
\end{equation}
the conditional probabilities are learned by the model function $f$. This training process is often called teacher forcing. The model can be then used to autoregressively generate the elements step-by-step:
\begin{equation}
    x_i = \text{argmax}~p(x | x_{<i}).
\end{equation}
State-of-the-art generative models such as GPTs \cite{gpt2, gpt3} are autoregressive generators with Transformer networks as the model function. They achieve great success in generating natural languages and images. In 3D generation, several studies have been conducted based on autoregressive models. In this section, we discuss some notable examples of employing autoregressive models for 3D generation.

PointGrow \cite{pointgrow} generates point clouds using an autoregressive network with self-attention context awareness operations in a point-by-point manner. Given its previously generated points, PointGrow reforms the points by axes and passes them into three branches. Each branch takes the inputs to predict a coordinate value of one axis. The model can also condition an embedding vector to generate point clouds, which can be a class category or an image. Inspired by the network from PointGrow, PolyGen \cite{polygen} generates 3D meshes with two transformer-based networks, one for vertices and one for faces. The vertex transformer autoregressively generates the next vertex coordinate based on previous vertices. The face transformer takes all the output vertices as context to generate faces. PolyGen can condition on a context of object classes or images, which are cross-attended by the transformer networks.


Recently, AutoSDF \cite{autosdf} generates 3D shapes represented by volumetric truncated-signed distance function (T-SDF). AutoSDF learns a quantized codebook regarding local regions of T-SDFs using VQ-VAE. The shapes are then presented by the codebook tokens and learned by a transformer-based network in a non-sequential autoregressive manner. In detail, given previous tokens at arbitrary locations and a query location, the network predicts the token that is queried. AutoSDF is capable of completing shapes and generating shapes based on images or text. Concurrently with AutoSDF, ShapeFormer \cite{shapeformer} generates surfaces of 3D shapes based on incomplete and noisy point clouds. A compact 3D representation called vector quantized deep implicit function (VQDIF) is used to represent shapes using a feature sequence of discrete variables. ShapeFormer first encodes an input point cloud into a partial feature sequence. It then uses a transformer-based network to autoregressively sample out the complete sequence. Finally, it decodes the sequence to a deep implicit function from which the complete object surface can be extracted. Instead of learning in 3D volumetric space, Luo~\etal proposes an improved auto-regressive model (ImAM) to learn discrete representation in a one-dimensional space to enhance the efficient learning of 3D shape generation. The method first encodes 3D shapes of volumetric grids into three axis-aligned planes. It uses a coupling network to further project the planes into a latent vector, where vector quantization is performed for discrete tokens. ImAM adopts a vanilla transformer to autoregressively learn the tokens with tractable orders. The generated tokens are decoded to occupancy values via a network by sampling spatial locations. ImAM can switch from unconditional generation to conditional generation by concatenating various conditions, such as point clouds, categories, and images.



\subsubsection{Variational Autoencoders}

Variational autoencoders (VAEs) \cite{vae} are probabilistic generative models that consist of two neural network components: the encoder and decoder. The encoder maps the input data point to a latent space that corresponds to the parameters of a variational distribution. In this way, the encoder can produce multiple different samples that all come from the same distribution. The decoder maps from the latent space to the input space, to produce or generate data points. Both networks are typically trained together with the usage of the reparameterization trick, although the variance of the noise model can be learned separately. VAEs have also been explored in 3D generation~\cite{setvae, tmnet, sdmnet, brock2016generative, nerfvae}. 

Brock~\etal~trains variational autoencoders directly for voxels using 3D ConvNet, while SDM-Net \cite{sdmnet} focuses on the generation of structured meshes composed of deformable parts. The method uses one VAE network to model parts and another to model the whole object. The follow-up work TM-Net~\cite{tmnet} could generate texture maps of meshes in a part-aware manner. Other representations like point clouds~\cite{setvae} and NeRFs~\cite{nerfvae} are also explored in variational autoencoders. Owing to the reconstruction-focused objective of VAEs, their training is considerably more stable than that of GANs. However, VAEs tend to produce more blurred results compared to GANs.




\subsubsection{Normalizing Flows}

Normalizing flow models consist of a series of invertible transformations that map a simple distribution, such as Gaussian, to a target distribution, which represents the data to generation. These transformations are carefully designed to be differentiable and invertible, allowing one to compute the likelihood of the data under the model and optimize the model parameters using gradient-based optimization techniques. 

In 3D generation, PointFlow \cite{pointflow} learns a distribution of shapes and a distribution of points using continuous normalizing flows. This approach allows for the sampling of shapes, followed by the sampling of an arbitrary number of points from a given shape. Discrete PointFlow (DPF) network \cite{dpf} improves PointFlow by replacing continuous normalizing flows with discrete normalizing flows, which reduces the training and sampling time. SoftFlow \cite{softflow} is a framework for training normalizing flows on the manifold. It estimates a conditional distribution of the perturbed input data instead of learning the data distribution directly. SoftFlow alleviates the difficulty of forming thin structures for flow-based models.
\subsection{Optimization-based Generation}
\label{sec:opt_gen}
Optimization-based generation is employed to generate 3D models using runtime optimization. These methods usually leverage pre-trained multimodal networks to optimize 3D models based on user-specified prompts. The key lies in achieving alignment between the given prompts and the generated content while maintaining high fidelity and diversity. In this section, we primarily examine optimization-based generation methods that use texts and images, based on the types of prompts provided by users.

\begin{figure}[t]
\begin{center}
   \includegraphics[width=1.0\linewidth]{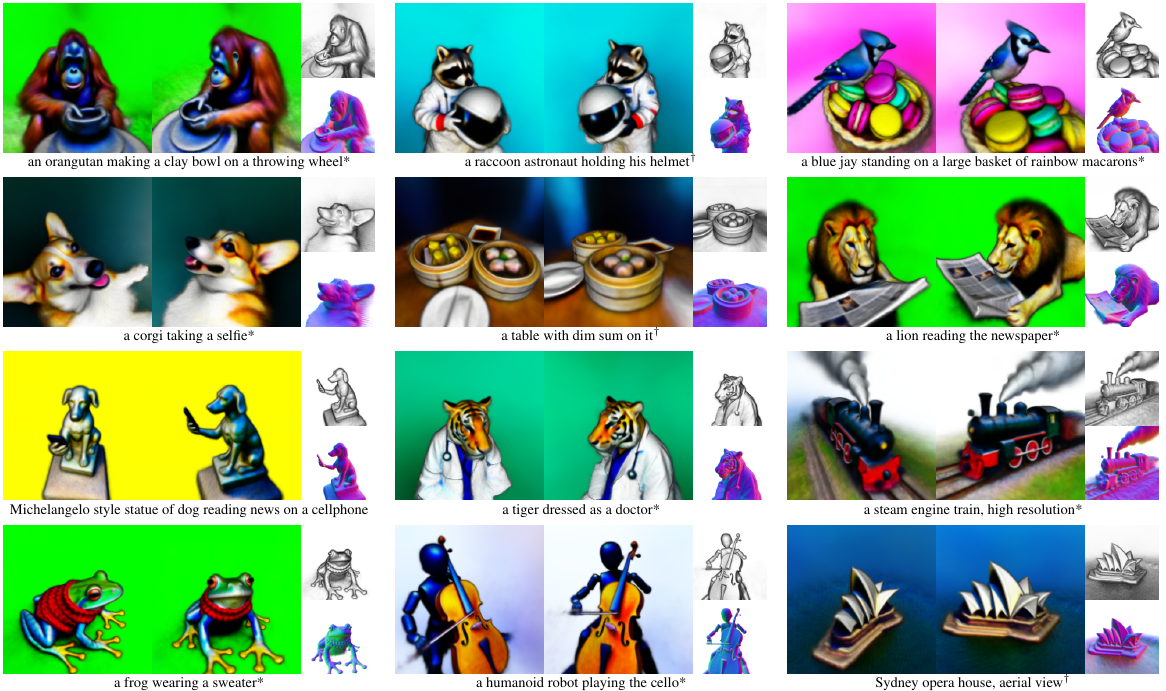}
\end{center}
\caption{Results of text-guided 3D generation by DreamFusion~\cite{poole2022dreamfusion} using SDS loss. $\ast$ denotes a DSLR photo, $\dag$ denotes a zoomed out DSLR photo.}
\label{fig:optimization}
\end{figure}

\subsubsection{Text-to-3D}
Language serves as the primary means of human communication and describing scenes, and researchers are dedicated to exploring the potential of text-based generation methods. These methods typically align the text with the images obtained through the differentiable rendering techniques, thereby guiding the generation of 3D content based on the text prompts. Given a fixed surface, TANGO~\cite{lei2022tango} uses CLIP~\cite{radford2021clip} to supervise differentiable physical-based rendering (PBR) images and obtain texture maps that align with the specified text prompt. Inspired by the success of NeRF~\cite{mildenhall2020nerf} and diffusion models in modeling 3D static scenes and text-to-image tasks respectively, DreamFusion~\cite{poole2022dreamfusion} (as shown in Fig.~\ref{fig:optimization}) combines the volumetric representation used in NeRF with the proposed Score Distillation Sampling (SDS) loss to achieve high-fidelity 3D content generation. SDS loss converts rendering error minimization into probability density distillation and enables 2D diffusion priors to optimize 3D representations (\eg, volumetric representation and triangle mesh) via image parameterization (\eg, differentiable rendering). As a concurrent work concurrent with SDS, Score Jacobian Chaining (SJC)~\cite{wang2023score} interprets predictions from pre-trained diffusion models as a score function of the data log-likelihood, similarly enabling 2D diffusion priors to optimize 3D representations via score matching. Based on DreamFusion, Magic3D~\cite{lin2023magic3d} introduces a coarse-to-fine manner and extracts the underlying geometry of the volume as a mesh. It then combines differentiable neural rendering and SDS to refine the extracted mesh. Magic3D is capable of exporting high-quality textured meshes and seamlessly embedding them into the traditional graphics pipeline. Also as a two-stage method, Fantasia3D further combines DMTet~\cite{shen2021deep} and SDS in the first geometry modeling stage to explicitly optimize surface. In the second stage, it introduces the PBR material model and disentangle texture and environment illumination. ProlificDreamer~\cite{wang2023prolificdreamer} presents variational score distillation (VSD) to boost text-to-3D generation. VSD adopts particles to model the distribution of 3D scenes and derive a gradient-based optimization scheme from the Wasserstein gradient flow, narrowing the gap between the rendering results distribution of the modeling distribution and pre-trained diffusion distribution. Benefiting from the optimization of scene distribution rather than a single scene, VSD overcomes the over-saturated and over-smoothed results produced by SDS and improves diversities. MVDream~\cite{shi2023mvdream} further fine-tunes a multi-view diffusion model and introduces multi-view consistent 3D priors, overcoming multi-face and content-drift problems. Text-to-3D has garnered significant attention recently, in addition to these, many other methods~\cite{zhu2023hifa, li2023sweetdreamer, metzer2023latent} have been proposed in this field.

\begin{table}[tb]
    \centering
    \small
    \caption{Quantitative comparison of image-to-3D methods on surface reconstruction. We summarize the Chamfer distance and volume IoU as the metrics to evaluate the quality of surface reconstruction.}
    \setlength{\tabcolsep}{12pt} 
    \begin{tabular}{@{}lcc@{}}
        \toprule
        Method & Chamfer Distance  $\downarrow$ & Volume IoU  $\uparrow$\\
        \midrule
        RealFusion~\cite{melas2023realfusion} & 0.0819 & 0.2741 \\
        Magic123 \cite{qian2023magic123} & 0.0516 & 0.4528 \\
        Make-it-3D \cite{tang2023make} & 0.0732 & 0.2937 \\
        One-2-3-45 \cite{liu2023one} & 0.0629 & 0.4086 \\
        Point-E \cite{nichol2022point} & 0.0426 & 0.2875 \\
        Shap-E \cite{jun2023shap} & 0.0436 & 0.3584 \\
        Zero-1-to-3~\cite{liu2023zero} & 0.0339 & 0.5035 \\
        SyncDreamer \cite{liu2023syncdreamer} & 0.0261 & 0.5421 \\
        \bottomrule
    \end{tabular}
    \label{tab:comparison_img_geo}
\end{table}

\begin{table}[]
    \centering
    \small
    \setlength{\tabcolsep}{8pt} 
    \caption{Quantitative comparison of image-to-3D methods on novel view synthesis. We report the CLIP-Similarity, PSNR, and LPIPS as the metrics to evaluate the quality of view synthesis. }
    \begin{tabular}{@{}lccc@{}}
        \toprule
        Method & CLIP-Similarity  $\uparrow$ & PSNR $\uparrow$ & LPIPS $\downarrow$ \\
        \midrule
        RealFusion~\cite{melas2023realfusion} & 0.735 & 20.216 & 0.197 \\ 
        Magic123 \cite{qian2023magic123} & 0.747 & 25.637 & 0.062 \\
        Make-it-3D \cite{tang2023make} & 0.839 & 20.010 & 0.119 \\ 
        One-2-3-45 \cite{liu2023one} & 0.788 & 23.159 & 0.096 \\
        Zero-1-to-3~\cite{liu2023zero} & 0.759 & 25.386 & 0.068 \\ 
        SyncDreamer \cite{liu2023syncdreamer} & 0.837 & 25.896 & 0.059 \\
        \bottomrule
    \end{tabular}
    \label{tab:comparison_img_view}
\end{table}

\subsubsection{Image-to-3D}
As the primary way to describe the visual effects of scenes, images can more intuitively describe the details and appearance of scenes at a finer-grained than language. Recent works thus are motivated to explore the image-to-3D techniques, which reconstruct remarkable and high-fidelity 3D models from specified images. These methods strive to maintain the appearance of the specified images and optimized 3D contents while introducing reasonable geometric priors. Similar to the text-to-3D methods, several image-to-3D methods leverage the volumetric representation used in NeRF to represent the target 3D scenes, which natively introduces multi-view consistency. NeuralLift-360~\cite{xu2023neurallift} uses estimated monocular depth and CLIP-guided diffusion prior to regularizing the geometry and appearance optimization respectively, achieving lift of a single image to a 3D scene represented by a NeRF. RealFusion~\cite{melas2023realfusion} and NeRDi~\cite{deng2023nerdi} leverage textual inversion~\cite{gal2022image} to extract text embeddings to condition a pre-trained image diffusion model~\cite{rombach2022high}, and combine use the score distillation loss to optimize the volumetric representation. Based on Magic3D \cite{lin2023magic3d} that employs a coarse-to-fine framework as mentioned above, Magic123~\cite{qian2023magic123} additionally introduces 3D priors from a pre-trained viewpoint-conditioned diffusion model Zero-1-to-3~\cite{liu2023zero} in two optimization stage, yielding textured meshes that match the specified images. As another two-stage image-to-3D method, Make-it-3D~\cite{tang2023make} enhances texture and geometric structure in the fine stage, producing high-quality textured point clouds as final results. Subsequent works~\cite{sun2023dreamcraft3d, yu2023hifi} have been consistently proposed to enhance the previous results. Recently, 3D Gaussian Splatting (3DGS)~\cite{kerbl20233d} has emerged as a promising modeling as well as a real-time rendering technique. Based on 3DGS, DreamGaussian~\cite{tang2023dreamgaussian} presents an efficient two-stage framework for both text-driven and image-driven 3D generation. In the first stage, DreamGaussian leverages SDS loss (\ie 2D diffusion priors~\cite{liu2023zero} and CLIP-guided diffusion priors~\cite{poole2022dreamfusion}) to generate target objects represented by 3D Gaussians. Then DreamGaussian extracts textured mesh from the optimized 3D Gaussians by querying the local density and refines textures in the UV space. For a better understanding of readers to various image-to-3D methods, we evaluate the performance of some open-source state-of-the-art methods. Tab. \ref{tab:comparison_img_geo} shows the quantitative comparison of image-to-3D methods on surface reconstruction. We summarize the Chamfer distance and volume IoU as the metrics to evaluate the quality of surface reconstruction. Tab. \ref{tab:comparison_img_view} demonstrates the quantitative comparison of image-to-3D methods on novel view synthesis. We report the CLIP-Similarity, PSNR, and LPIPS as the metrics to evaluate the quality of view synthesis.



\subsection{Procedural Generation}
\label{sec:pro_gen}

Procedural generation is a term for techniques that create 3D models and textures from sets of rules. These techniques often rely on predefined rules, parameters, and mathematical functions to generate diverse and complex content, such as textures, terrains, levels, characters, and objects. One of the key advantages of procedural generation is their ability to efficiently create various shapes from a relatively small set of rules. In this section, we mainly survey four most used techniques: fractal geometry, L-Systems, noise functions and cellular automata.

A fractal \cite{mandelbrot1967long,mandelbrot1982fractal} is a geometric shape that exhibits detailed structure at arbitrarily small scales. A characteristic feature of many fractals is their similarity across different scales. This property of exhibiting recurring patterns at progressively smaller scales is referred to as self-similarity. A common application of fractal geometry is the creation of landscapes or surfaces. These are generated using a stochastic algorithm designed to produce fractal behavior that mimics the appearance of natural terrain. The resulting surface is not deterministic, but rather a random surface that exhibits fractal behavior.

An L-system \cite{lsystem}, or Lindenmayer system, is a type of formal grammar and parallel rewriting system. It comprises an alphabet of symbols that can be utilized to construct strings, a set of production rules that transform each symbol into a more complex string of symbols, a starting string for construction, and a mechanism for converting the produced strings into geometric structures. L-systems are used to create complex and realistic 3D models of natural objects like trees and plants. The string generated by the L-System can be interpreted as instructions for a ``turtle'' to move in 3D space. For example, certain characters might instruct the turtle to move forward, turn left or right, or push and pop positions and orientations onto a stack.

Noise functions, such as Perlin noise \cite{perlin1985image} and Simplex noise \cite{perlin2002improving}, are used to generate coherent random patterns that can be applied to create realistic textures and shapes in 3D objects. These functions can be combined and layered to create more complex patterns and are particularly useful in terrain generation, where they can be used to generate realistic landscapes with varying elevations, slopes, and features.

Cellular automata~\cite{von1951general,neumann1966theory,wolfram1983statistical} are a class of discrete computational models that consist of a grid of cells, each of which can be in one of a finite number of states. The state of each cell is determined by a set of rules based on the states of its neighboring cells. Cellular automata have been used in procedural generation to create various 3D objects and patterns, such as cave systems, mazes, and other structures with emergent properties.
\subsection{Generative Novel View Synthesis}
\label{sec:gnvs}

\begin{figure}[t]
\begin{center}
   \includegraphics[width=0.95\linewidth]{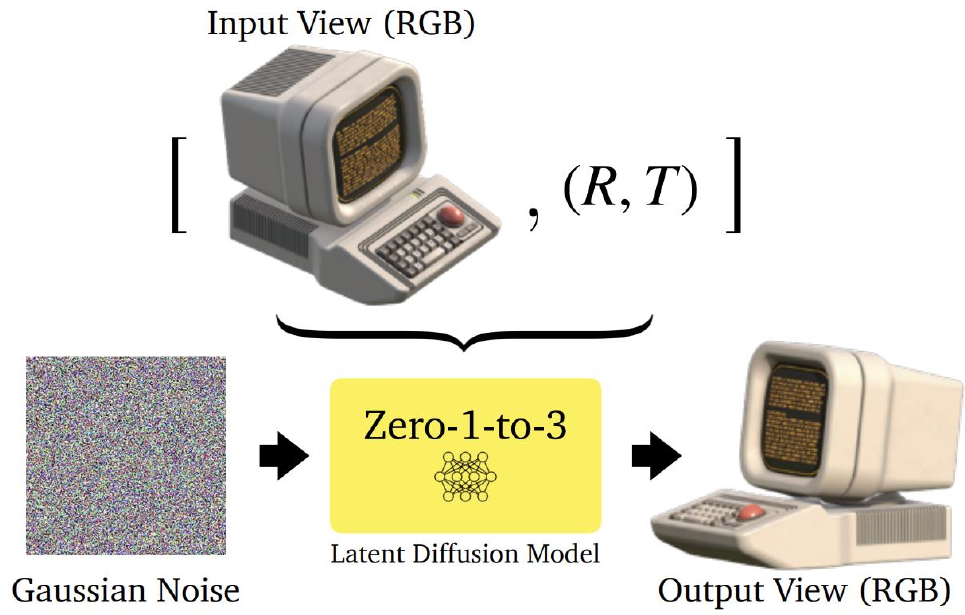}
\end{center}
\caption{Zero-1-to-3 proposes a viewpoint-conditioned image diffusion model to generate the novel view of the input image. By training on a large-scale dataset, it achieves a strong generalization ability to in-the-wild images.}
\label{fig:zero123}
\end{figure}

Recently, generative techniques have been utilized to tackle the challenge of novel view synthesis, particularly in predicting new views from a single input image. Compared to the conventional 3D generation methods, it does not explicitly utilize the 3D representation to enforce 3D consistency, instead, it usually employs a 3D-aware method by conditioning 3D information. In the field of novel view synthesis, a widely studied technical route will be regression-based methods~\cite{yu2021pixelnerf, wang2021ibrnet, chen2021mvsnerf}. Different from them, generative novel view synthesis focuses more on generating new content rather than regressing the scenes from a few input images, which typically involves long-range view extrapolation. 

\begin{table*}[t]
\centering
\small
\caption{Selected datasets commonly used for 3D generation.}
\begin{tabular}{@{}lcccc@{}}
	\toprule
        Dataset & Type & Year & Samples & Category \\
        \midrule
	ShapeNet~\cite{chang2015shapenet} & 3D data & 2015 & 51K & objects \\
        Thingi10K~\cite{zhou2016thingi10k} & 3D data & 2016 & 10K & objects \\
        3D-Future~\cite{fu20213d} & 3D data & 2020 & 10K & furniture \\
        GSO~\cite{downs2022google} & 3D Data & 2022 & 1K & household items \\
        Objaverse~\cite{deitke2023objaverse}  & 3D data & 2022 & 800K & objects \\
        OmniObject3D~\cite{wu2023omniobject3d} & 3D data & 2023 & 6K & objects \\
	
	Objaverse-XL~\cite{deitke2023objaversexl} & 3D Data & 2023 & 10.2M & objects \\
	\cmidrule(lr){1-5}
        ScanNet~\cite{dai2017scannet}  & multi-view images & 2017 & 1.5K (2.5M images) &  indoor scenes \\
        CO3D~\cite{reizenstein2021common}  & multi-view images & 2021 & 19K (1.5M images) & objects \\
	MVImgNet~\cite{yu2023mvimgnet}  & multi-view images & 2023 & 219K (6.5M images) & objects \\
        \cmidrule(lr){1-5}
	DeepFashion~\cite{liu2016deepfashion} & single-view images & 2016 & 800K & clothes \\
        FFHQ~\cite{karras2019style} & single-view images & 2018 & 70K & human faces \\
        AFHQ~\cite{choi2020stargan} & single-view images & 2019 & 15K & animal faces \\
        SHHQ~\cite{fu2022stylegan} & single-view images & 2022 & 40K & human bodies \\
	\bottomrule
\end{tabular}
\label{tab:datasets}
\end{table*}

With the development of image synthesis methods, significant progress has been made in generative novel view synthesis. Recently, 2D diffusion models have transformed image synthesis and therefore are also utilized in generative novel view synthesis~\cite{watson2022novel, tseng2023consistent, liu2023zero, chan2023generative, tewari2023diffusion, yoo2023dreamsparse}. Among these methods, 3DiM~\cite{watson2022novel} first introduces a geometry-free image-to-image diffusion model for novel view synthesis, taking the camera pose as the condition. Tseng~\etal~\cite{tseng2023consistent} designs epipolar attention layers to inject camera parameters into the pose-guided diffusion model for consistent view synthesis from a single input image. Zero-1-to-3~\cite{liu2023zero} (as shown in Fig.~\ref{fig:zero123}) demonstrates the learning of the camera viewpoint in large-scale diffusion models for zero-shot novel view synthesis. ~\cite{chan2023generative, tewari2023diffusion, yoo2023dreamsparse} condition 2D diffusion models on pixel-aligned features extracted from input views to extend them to be 3D-aware. However, generating multiview-consistent images remains a challenging problem. To ensure consistent generation,~\cite{lin2023consistent123, liu2023syncdreamer, shi2023zero123++, long2023wonder3d} propose a multi-view diffusion model that could synthesize multi-view images simultaneously to consider the information between different views, which achieve more consistent results compared to the single view synthesis model like Zero-1-to-3~\cite{liu2023zero}. 

Prior to that, the transformer which is a sequence-to-sequence model originally proposed in natural language processing, uses a multi-head attention mechanism to gather information from different positions and brings lots of attention in the vision community. Many tasks achieve state-of-the-art performance using the attention mechanism from the transformer including generative novel view synthesis~\cite{rombach2021geometry, sajjadi2022scene, kulhanek2022viewformer}. Specifically, Geometry-free View Synthesis~\cite{rombach2021geometry} learns the discrete representation vis VQGAN to obtain an abstract latent space for training transformers. While ViewFormer~\cite{kulhanek2022viewformer} also uses a two-stage training consisting of a Vector Quantized Variational Autoencoder (VQ-VAE) codebook and a transformer model. And~\cite{sajjadi2022scene} employs an encoder-decoder model based on transformers to learn an implicit representation.

On the other hand, generative adversarial networks could produce high-quality results in image synthesis and consequently are applied to novel view synthesis~\cite{wiles2020synsin, koh2021pathdreamer, rockwell2021pixelsynth, liu2021infinite, li2022infinitenature}. Some methods~\cite{wiles2020synsin, koh2021pathdreamer, rockwell2021pixelsynth} maintain a 3D point cloud as the representation, which could be projected onto novel views followed by a GAN to hallucinate the missing regions and synthesize the output image. While~\cite{liu2021infinite} and ~\cite{li2022infinitenature} focus on long-range view generation from a single view with adversarial training. At an earlier stage of deep learning methods when the auto-encoders and variational autoencoders begin to be explored, it is also used to synthesize the novel views~\cite{kulkarni2015deep, zhou2016view, tatarchenko2016multi, chen2019monocular}.

In summary, generative novel view synthesis can be regarded as a subset of image synthesis techniques and continues to evolve alongside advancements in image synthesis methods. Besides the generative models typically included, determining how to integrate information from the input view as a condition for synthesizing the novel view is the primary issue these methods are concerned with.

\section{Datasets for 3D Generation} 
\label{sec:datasets}

With the rapid development of technology, the ways of data acquisition and storage become more feasible and affordable, resulting in an exponential increase in the amount of available data. As data accumulates, the paradigm for problem-solving gradually shifts from data-driven to model-driven approaches, which in turn contributes to the growth of "Big Data" and "AIGC". Nowadays, data plays a crucial role in ensuring the success of algorithms. A well-curated dataset can significantly enhance a model's robustness and performance. On the contrary, noisy and flawed data may cause model bias that requires considerable effort in algorithm design to rectify. In this section, we will go over the common data used for 3D generation. Depending on the methods employed, it usually includes 3D data (Section~\ref{sec:3d_data}), multi-view image data (Section~\ref{sec:multi_view_data}), and single-view image data (Section~\ref{sec:single_view_data}), which are also summarized in Tab.~\ref{tab:datasets}.

\subsection{Learning from 3D Data}
\label{sec:3d_data}
3D data could be collected by RGB-D sensors and other technology for scanning and reconstruction. Apart from 3D generation, 3D data is also widely used for other tasks like helping improve classical 2D vision task performance by data synthesis, environment simulation for training embodied AI agents, 3D object understanding, etc. One popular and frequently used 3D model database in the early stage is The Princeton Shape Benchmark~\cite{shilane2004princeton}. It contains about 1800 polygonal models collected from the World Wide Web. While~\cite{kasper2012kit} constructs a special rig that contains a 3D digitizer, a turntable, and a pair of cameras mounted on a sled that can move along a bent rail to capture the kit object models database. To evaluate the algorithms to detect and estimate the objects in the image given 3D models,~\cite{lim2013parsing} introduces a dataset of 3D IKEA models obtained from Google Warehouse. Some 3D model databases are presented for tasks like robotic manipulation~\cite{calli2015benchmarking, morrison2020egad}, 3D shape retrieval~\cite{li2014shrec}, 3D shape modeling from a single image~\cite{sun2018pix3d}. BigBIRD~\cite{singh2014bigbird} presents a large-scale dataset of 3D object instances that also includes multi-view images and depths, camera pose information, and segmented objects for each image.

However, those datasets are very small and only contain hundreds or thousands of objects. Collecting, organizing, and labeling larger datasets in computer vision and graphics communities is needed for data-driven methods of 3D content. To address this, ShapeNet~\cite{chang2015shapenet} is introduced to build a large-scale repository of 3D CAD models of objects. The core of ShapeNet covers 55 common object categories with about 51,300 models that are manually verified category and alignment annotations. Thingi10K~\cite{zhou2016thingi10k} collects 10,000 3D printing models from an online repository Thingiverse. While PhotoShape~\cite{park2018photoshape} produces 11,000 photorealistic, relightable 3D shapes based on online data. Other datasets such as 3D-Future~\cite{fu20213d}, ABO~\cite{collins2022abo}, GSO~\cite{downs2022google} and OmniObject3D~\cite{wu2023omniobject3d} try to improve the texture quality but only contain thousands of models. Recently, Objaverse~\cite{deitke2023objaverse} presents a large-scale corpus of 3D objects that contains over 800K 3D assets for research in the field of AI and makes a step toward a large-scale 3D dataset. Objaverse-XL~\cite{deitke2023objaversexl} further extends Objaverse to a larger 3D dataset of 10.2M unique objects from a diverse set of sources. These large-scale 3D datasets have the potential to facilitate large-scale training and boost the performance of 3D generation.

\subsection{Learning from Multi-view Images}
\label{sec:multi_view_data}
3D objects have been traditionally created through manual 3D modeling, object scanning, conversion of CAD models, or combinations of these techniques~\cite{downs2022google}. These techniques may only produce synthetic data or real-world data of specific objects with limited reconstruction accuracy. Therefore, some datasets directly provide multi-view images in the wild which are also widely used in many 3D generation methods. ScanNet~\cite{dai2017scannet} introduces an RGB-D video dataset containing 2.5M views in 1513 scenes and Objectron~\cite{ahmadyan2021objectron} contains object-centric short videos and includes 4
million images in 14,819 annotated videos, of which only a limited number cover the full 360 degrees. CO3D~\cite{reizenstein2021common} extends the dataset from~\cite{henzler2021unsupervised} and increases the size to nearly 19,000 videos capturing objects from 50 MS-COCO categories, which has been widely used in the training and evaluations of novel view synthesis and 3D generation or reconstruction methods. Recently, MVImgNet~\cite{yu2023mvimgnet} presents a large-scale dataset of multi-view images that collects 6.5 million frames from 219,188 videos by shooting videos of real-world objects in human daily life. Other lines of work provide the multi-view dataset in small-scale RGB-D videos~\cite{lai2011large, shrestha2022real, chao2023fewsol} compared with these works, large-scale synthetic videos~\cite{tremblay2022rtmv}, or egocentric videos~\cite{zhu2023egoobjects}. A large-scale dataset is still a remarkable trend for deep learning methods, especially for generation tasks.

\subsection{Learning from Single-view Images}
\label{sec:single_view_data}
3D generation methods usually rely on multi-view images or 3D ground truth to supervise the reconstruction and generation of 3D representation. Synthesizing high-quality multi-view images or 3D shapes using only collections of single-view images is a challenging problem. Benefiting from the unsupervised training of generative adversarial networks, 3D-aware GANs are introduced that could learn 3D representations in an unsupervised way from natural images. Therefore, several single-view image datasets are proposed and commonly used for these 3D generation methods. Although many large-scale image datasets have been presented for 2D generation, it is hard to directly use them for 3D generation due to the high uncertainty of this problem. Normally, these image datasets only contain a specific category or domain. FFHQ~\cite{karras2019style}, a real-world human face dataset consisting of 70,000 high-quality images at $1024^2$ resolution, and AFHQ~\cite{choi2020stargan}, an animal face dataset consisting of 15,000 high-quality images at $512^2$ resolution, are introduced for 2D image synthesis and used a lot for 3D generation based on 3D-aware GANs. In the domain of the human body, SHHQ~\cite{fu2022stylegan} and DeepFashion~\cite{liu2016deepfashion} have been adopted for 3D human generation. In terms of objects, many methods~\cite{liao2020towards, gadelha20173d, henzler2019escaping, zhu2018visual, wu2016learning} render synthetic single-view datasets using several major object categories of ShapeNet. While GRAF~\cite{schwarz2020graf} renders 150k Chairs from Photoshapes~\cite{park2018photoshape}. Moreover, CelebA~\cite{liu2015deep} and Cats~\cite{zhang2008cat} datasets are also commonly used to train the models like HoloGAN~\cite{nguyen2019hologan} and pi-GAN~\cite{chan2021pi}. Since the single-view images are easy to obtain, these methods could collect their own dataset for the tasks.

\section{Applications} \label{sec:app}
\label{sec:appli}
In this section, we introduce various 3D generation tasks (Sec.~\ref{sec:app:human_gen}-\ref{sec:app:generic_gen}) and closely related 3D editing tasks (Sec.~\ref{sec:app:editing}).
The generation tasks are divided into three categories, including 3D human generation (Sec.~\ref{sec:app:human_gen}), 3D face generation (Sec.~\ref{sec:app:face_gen}), and generic object and scene generation (Sec.~\ref{sec:app:generic_gen}).

\begin{table}[t]
\centering
\small
\setlength { \tabcolsep } {11pt}
\caption{Recent 3D human generation techniques and their corresponding input-output formats.}
\begin{tabular}{@{}lccc@{}}
\toprule
    Methods & Input Condition & \thead{Output Texture} \\
    \midrule
    ICON~\cite{xiu2022icon}  & Single-Image & \XSolidBrush\\
    ECON~\cite{xiu2023econ}  & Single-Image & \XSolidBrush\\
    gDNA~\cite{chen2022gdna}  & Latent & \XSolidBrush\\
    Chupa~\cite{kim2023chupa}  & Text/Latent & \XSolidBrush\\
    ELICIT~\cite{huang2023one} & Single-Image & \Checkmark \\
    TeCH~\cite{huang2023tech} & Single-Image & \Checkmark \\
    Get3DHuman~\cite{xiong2023get3dhuman} & Latent & \Checkmark \\
    EVA3D~\cite{hong2022eva3d} & Latent & \Checkmark \\
    AvatarCraft~\cite{jiang2023avatarcraft} & Text & \Checkmark \\
    DreamHuman~\cite{kolotouros2023dreamhuman} & Text & \Checkmark  \\
    TADA~\cite{liao2023tada}  & Text & \Checkmark \\
\bottomrule
\end{tabular}
\label{tab:app_human}
\end{table}

\begin{figure}[t]
\begin{center}
   \includegraphics[width=1.0\linewidth]{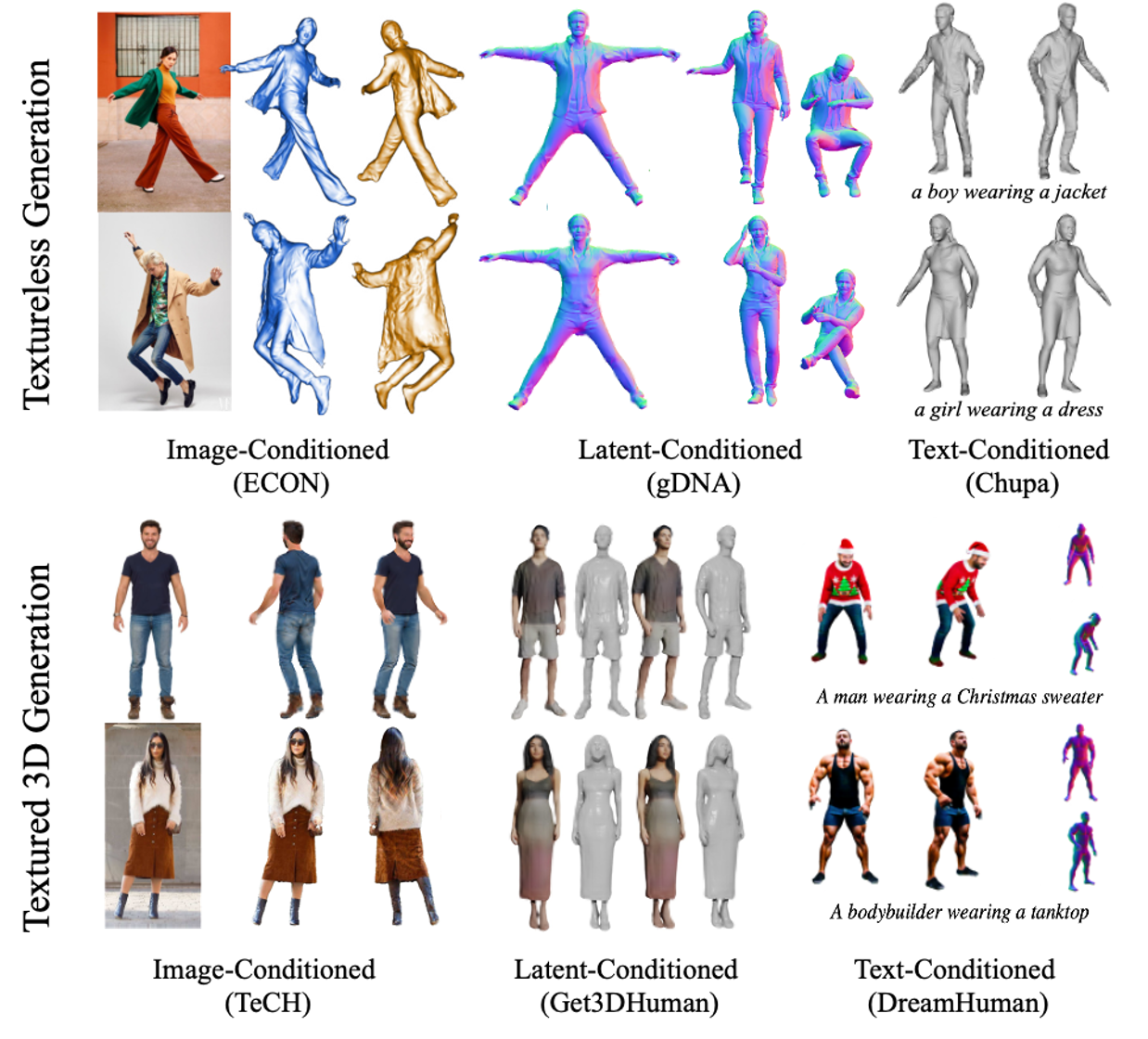}
\end{center}
\caption{Examples of 3D human generation methods. 3D generation results source from ECON~\cite{xiu2023econ}, gDNA~\cite{chen2022gdna}, Chupa~\cite{kim2023chupa}, TeCH~\cite{huang2023tech}, Get3DHuman~\cite{xiong2023get3dhuman}, and DreamHuman~\cite{kolotouros2023dreamhuman}.}
\label{fig:app_results_human}
\end{figure}

\subsection{3D Human Generation}
\label{sec:app:human_gen}

    
With the emergence of the metaverse and the advancements in virtual 3D social interaction, the field of 3D human digitization and generation has gained significant attention in recent years. 
Different from general 3D generation methods that focus on category-free rigid objects with sample geometric structures \cite{poole2022dreamfusion, lorraine2023att3d}, most 3D human generation methods aim to tackle the complexities of articulated pose changes and intricate geometric details of clothing. 
Tab.~\ref{tab:app_human} presents a compilation of notable 3D human body generation methods in recent years, organized according to the input conditions and the output format of the generated 3D human bodies. Some results of these methods are shown in Fig.~\ref{fig:app_results_human}.
Specifically, in terms of the input condition, current 3D human body generation methods can be categorized based on the driving factors including latent features randomly sampled from a pre-defined latent space \cite{ma2020learning, chen2022gdna, hong2022eva3d}, a single reference image \cite{alldieck2019tex2shape, corona2021smplicit, xiu2023econ, huang2023tech, zhang2023humanref}, or text prompts \cite{kim2023chupa, jiang2023avatarcraft, kolotouros2023dreamhuman, liao2023tada}. 
According to the form of the final output, these methods can be classified into two categories: textureless shape generation~\cite{alldieck2019tex2shape, xiu2022icon, xiu2023econ, chen2022gdna, ma2020learning, corona2021smplicit, kim2023chupa} and textured body generation~\cite{alldieck2022photorealistic, liao2023tada, huang2023one, kolotouros2023dreamhuman, xiong2023get3dhuman, huang2023tech, zhang2023humanref}. 
While the latter focuses on generating fully textured 3D clothed humans, the former aims to obtain textureless body geometry with realistic details.

In terms of textureless shape generation, early works \cite{choutas2020monocular, osman2020star, li2021hybrik} attempt to predict SMPL parameters from the input image and infer a skinned SMPL mesh as the generated 3D representation of the target human. 
Nevertheless, such skinned body representation fails to represent the geometry of clothes. 
To overcome this issue, \cite{alldieck2019tex2shape, xiu2022icon, xiu2023econ} leverage a pre-trained neural network to infer the normal information and combine the skinned SMPL mesh to deduce a clothed full-body geometry with details. 
In contrast to such methods, which require reference images as input, CAPE \cite{ma2020learning} proposes a generative 3D mesh model conditioned on latents of SMPL pose and clothing type to form the clothing deformation from the SMPL body. 
gDNA \cite{chen2022gdna} introduces a generation framework conditioned on latent codes of shape and surface details to learn the underlying statistics of 3D clothing details from scanned human datasets via an adversarial loss. 
Different from the previous methods that generate an integrated 3D clothed human body geometry, SMPLicit~\cite{corona2021smplicit} adopts an implicit model conditioned on shape and pose parameters to individually generate diverse 3D clothes. By combining the SMPL body and associated generated 3D clothes, SMPLicit enables to produce 3D clothed human shapes. 
To further improve the quality of the generated human shape, Chupa \cite{kim2023chupa} introduces diffusion models to generate realistic human geometry and decompose the 3D generation task into 2D normal map generation and normal map-based 3D reconstruction.

Although these methods achieve the generation of detailed clothed human shapes, their application prospects are greatly restricted due to the lack of texture-generation capabilities. 
To generate textured clothed 3D human, lots of attempts have been made in previous work, including methods conditioned on latent codes \cite{grigorev2021stylepeople, bergman2022generative, zhang2022avatargen, noguchi2022unsupervised, jiang2023humangen, yang20223dhumangan, xiong2023get3dhuman, chen2023veri3d, hong2022eva3d, xu2023efficient, abdal2023gaussian}, 
single images \cite{saito2019pifu, zheng2021pamir, alldieck2022photorealistic, choi2022mononhr, gao2023contex, huang2023one, yang2023d, hu2023sherf, albahar2023single, huang2023tech, zhang2023humanref}, 
and text prompts \cite{hong2022avatarclip, jiang2023avatarcraft, cao2023dreamavatar, huang2023dreamwaltz, kolotouros2023dreamhuman, zhang2023avatarverse, liao2023tada, huang2023humannorm, zhang2023avatarstudio, liu2023humangaussian}.
Most latent-conditioned methods employ adversarial losses to restrict their latent space and generate 3D human bodies within the relevant domain of the training dataset. For example, StylePeople \cite{grigorev2021stylepeople} combines StyleGAN~\cite{karras2020analyzing} and neural rendering to design a joint generation framework trained in an adversarial fashion on the full-body image datasets. Furthermore, GNARF~\cite{bergman2022generative} and AvatarGen~\cite{zhang2022avatargen} employ tri-planes as the 3D representation and replace the neural rendering with volume rendering to enhance the view-consistency of rendered results. 
To improve editability, Get3DHuman~\cite{xiong2023get3dhuman} divides the human body generation framework into shape and texture branches respectively conditioned on shape and texture latent codes, achieving re-texturing. 
EVA3D~\cite{hong2022eva3d} divides the generated human body into local parts to achieve controllable human poses. 

As text-to-image models~\cite{radford2021learning, rombach2022high, saharia2022photorealistic} continue to advance rapidly, the field of text-to-3D has also reached its pinnacle of development. 
For the text-driven human generation, existing methods inject priors from pre-trained text-to-image models into the 3D human generation framework to achieve text-driven textured human generation, such as AvatarCLIP \cite{hong2022avatarclip}, AvatarCraft \cite{jiang2023avatarcraft}, DreamHuman \cite{kolotouros2023dreamhuman}, and TADA \cite{liao2023tada}. 
Indeed, text-driven human generation methods effectively address the challenge of limited 3D training data and significantly enhance the generation capabilities of 3D human assets. 
Nevertheless, in contrast to the generation of unseen 3D humans, it is also significant to generate a 3D human body from a specified single image in real-life applications. 
In terms of single-image-conditioned 3D human generation methods, producing generated results with textures and geometries aligned with the input reference image is widely studied. To this end, PIFu \cite{saito2019pifu}, PaMIR \cite{zheng2021pamir}, and PHORHUM \cite{alldieck2022photorealistic} propose learning-based 3D generators trained on scanned human datasets to infer human body geometry and texture from input images. 
However, their performance is constrained by the limitations of the training data. Consequently, they struggle to accurately infer detailed textures and fine geometry from in-the-wild input images, particularly in areas that are not directly visible in the input. To achieve data-free 3D human generation, ELICIT \cite{huang2023one}, Human-SGD \cite{albahar2023single}, TeCH \cite{huang2023tech}, and HumanRef \cite{zhang2023humanref} leverage priors of pre-trained CLIP~\cite{radford2021learning} or image diffusion models~\cite{rombach2022high,saharia2022photorealistic} to predict the geometry and texture based on the input reference image without the need for 3D datasets, and achieve impressive qualities in generated 3D clothed human.

\subsection{3D Face Generation}
\label{sec:app:face_gen}

\begin{figure}
\begin{center}
   \includegraphics[width=1.0\linewidth]{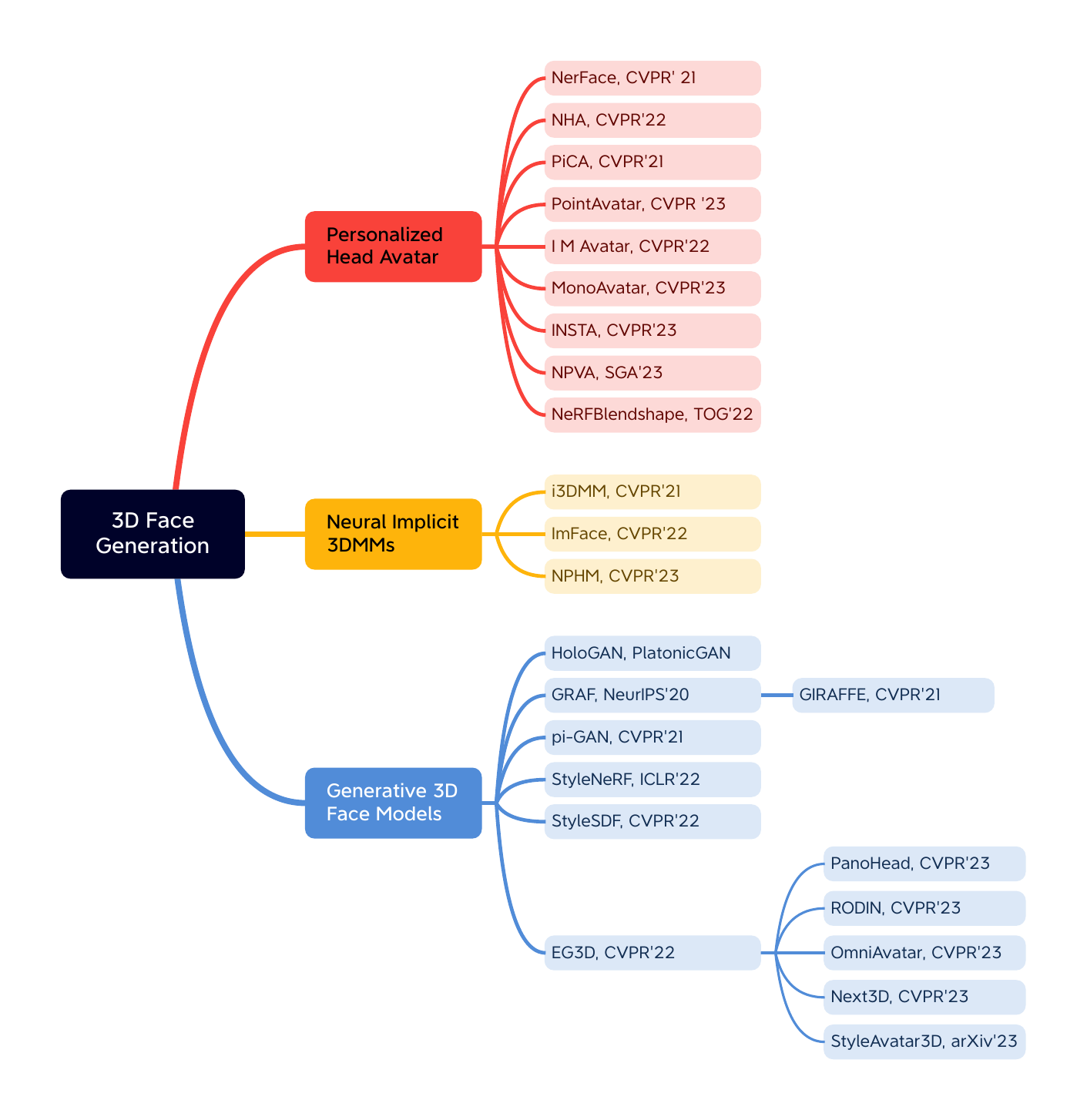}
\end{center}
\caption{
Representative applications and methods of 3D face generation.
}
\label{fig:app_face_tree}
\end{figure}

\begin{figure}[t]
\begin{center}
   \includegraphics[width=1.0\linewidth]{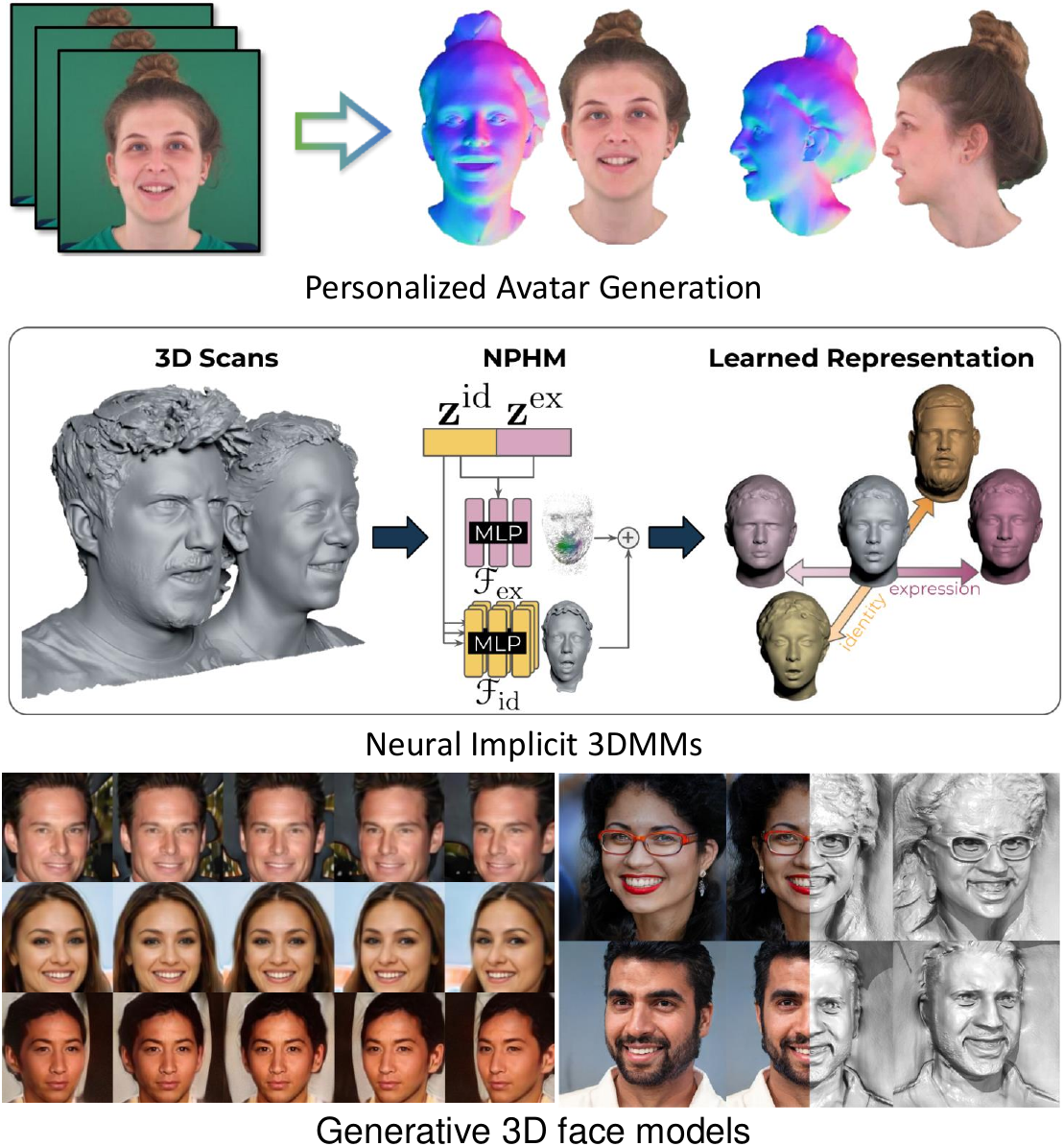}
\end{center}
\caption{
Representative 3D face generation tasks.
Images adapted from NHA~\cite{grassal2022nha}, NPHM~\cite{giebenhain2023nphm}, and EG3D~\cite{chan2022efficient}.
}
\label{fig:app_face_example}
\end{figure}

One essential characteristic of 3D face generation tasks is to generate high-quality human face images that can be viewed from different viewpoints.
Popular tasks can be loosely classified into three major categories, including personalized head avatar creation (e.g. 3D talking head generation), neural implicit 3D morphable models (3DMMs), and generative 3D face models, which are shown in Fig.~\ref{fig:app_face_tree} and Fig.~\ref{fig:app_face_example}.

\vspace{1mm}
\noindent\textbf{Personalized head avatar creation} aims at creating an animatable avatar that can be viewed from different viewpoints of the target person, which has broad applications such as talking head generation.
Most of the existing methods take as input a sequence of video frames (i.e. monocular video)~\cite{park2021nerfies,gafni2021dynamic,grassal2022nha,zheng2022avatar,zielonka2023instant,zheng2023pointavatar,bai2023monoavatar,gao2022nerfblendshape}.
Although convenient, the viewing angles of these avatars are limited in a relatively small range (i.e. near frontal) and their quality is not always satisfactory due to limited data.
In contrast, another stream of works~\cite{lombardi2018dam,ma2021pixel,lombardi2021mixture,wang2023neural,kirschstein2023nersemble} aims at creating a very high-quality digital human that can be viewed from larger angles (e.g. side view).
These methods usually require high-quality synchronized multi-view images under even illumination.
However, both streams rely heavily on implicit or hybrid neural representations and neural rendering techniques.
The quality and animation accuracy of the generated talking head video are usually measured with PSNR, SSIM, and LPIPS metrics.

\vspace{1mm}
\noindent\textbf{Neural implicit 3DMMs.} 
Traditional 3D morphable face models (3DMMs) assume a predefined template mesh (i.g. fixed topology) for the geometry and have explored various modeling methods including linear models (e.g. PCA-based 3DMMs) and non-linear models (e.g. network-based 3DMMs). 
A comprehensive survey of these methods has been discussed in \cite{egger20203dmm}.
Recently, thanks to the rapid advances in implicit neural representations (INRs), several neural implicit 3DMMs utilizing INRs for face modeling emerges~\cite{yenamandra2021i3dmm,zheng2022imface,giebenhain2023nphm} since continuous implicit neural representations do not face discretization error and can theoretically modeling infinite details.
Indeed, NPHM~\cite{giebenhain2023nphm} can generate more subtle expressions unseen in previous mesh-based 3DMMs.
What's more, neural implicit 3DMMs can potentially model hair better since the complexity of different hairstyles varies drastically, which imposes a great challenge for fixed topology mesh-based traditional 3DMMs.

\vspace{1mm}
\noindent\textbf{Generative 3D face models.}
One key difference from 2D generative face models (e.g. StyleGAN~\cite{karras2019style,karras2020analyzing}) is that 3D face models can synthesize multi-view consistent images (i.e. novel views) of the same target (identity and clothes).
Early attempts towards this direction include HoloGAN~\cite{nguyen2019hologan} and PlatonicGAN~\cite{henzler2019platonicgan}, which are both voxel-based methods and can only generate images in limited resolution.
Quickly, methods~\cite{schwarz2020graf,niemeyer2021giraffe,chan2021pigan,or2022stylesdf,gu2021stylenerf,chan2022efficient} utilizing neural radiance fields are proposed to increase the image resolution.
For example, EG3D~\cite{chan2022efficient} proposes a hybrid tri-plane representation, which strikes a good trade-off to effectively address the memory and rendering inefficiency faced by previous generative 3D GANs and can produce high-quality images with good multi-view consistency.

Thanks to the success of various 3D GANs, many downstream applications (e.g. editing, talking head generation) are enabled or become less data-hungry, including 3D consistent editing~\cite{sun2023next3d,sun2022fenerf,sun2022ide,lin2023sketchfacenerf,jiang2022nerffaceediting}, 3D talking head generation~\cite{bai2023high,xu2023omniavatar,wu2022anifacegan}, etc.

\begin{figure*}[t]
\begin{center}
   \includegraphics[width=1.0\linewidth]{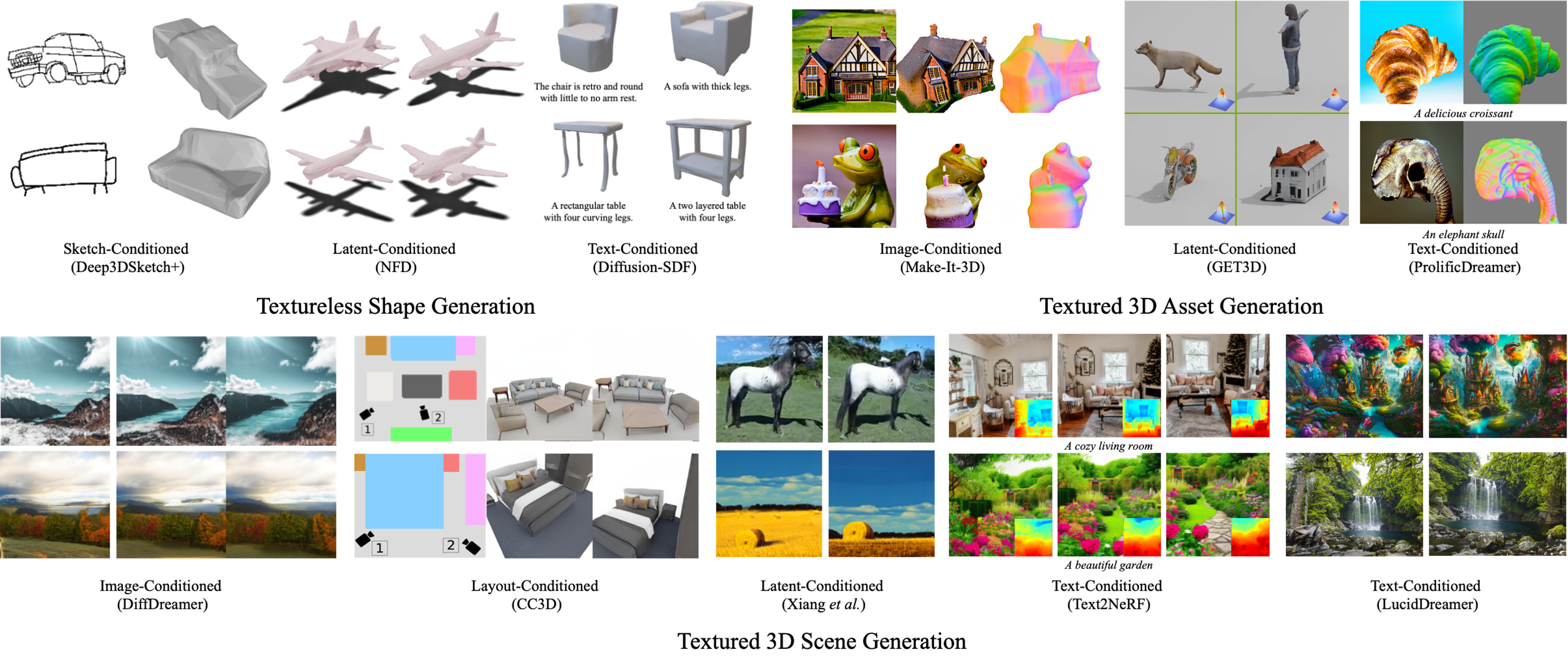}
\end{center}
\caption{Some examples of general scene generation methods. 3D generation results source from Deep3DSketch+~\cite{chen2023deep3dsketch+}, NFD~\cite{shue20233d}, Diffusion-SDF~\cite{li2023diffusion}, Make-It-3D~\cite{tang2023make}, GET3D~\cite{gao2022get3d}, ProlificDreamer~\cite{wang2023prolificdreamer}, DiffDreamer~\cite{cai2023diffdreamer}, CC3D~\cite{bahmani2023cc3d}, Xiang \emph{et al.}~\cite{xiang20233d}, Text2NeRF \cite{zhang2023text2nerf}, and LucidDreamer \cite{chung2023luciddreamer}.}
\label{fig:app_results_scene}
\end{figure*}

\begin{table}[t]
\centering
\small
    \setlength{\tabcolsep}{3pt} 
\caption{Applications of general scene generation methods.}
\begin{tabular}{@{}lccc@{}}
\toprule
    Methods & \thead{Type} & Condition & \thead{Texture \\ Generation} \\
    \midrule
    PVD~\cite{zhou20213d} & Object-Centered & Latent & \XSolidBrush \\
    NFD~\cite{shue20233d} & Object-Centered & Latent & \XSolidBrush \\
    Point-E~\cite{nichol2022point} & Object-Centered & Text & \XSolidBrush \\
    Diffusion-SDF~\cite{li2023diffusion} & Object-Centered & Text & \XSolidBrush \\
    Deep3DSketch+~\cite{chen2023deep3dsketch+} & Object-Centered & Sketch & \XSolidBrush \\
    Zero-1-to-3~\cite{liu2023zero} & Object-Centered & Single-Image & \Checkmark \\
    Make-It-3D~\cite{tang2023make} & Object-Centered & Single-Image & \Checkmark \\
    GET3D~\cite{gao2022get3d} & Object-Centered & Latent & \Checkmark \\
    EG3D~\cite{chan2022efficient} & Object-Centered & Latent & \Checkmark \\
    CLIP-Mesh~\cite{mohammad2022clip} & Object-Centered & Text & \Checkmark \\
    DreamFusion~\cite{poole2022dreamfusion} & Object-Centered & Text & \Checkmark \\
    ProlificDreamer~\cite{wang2023prolificdreamer} & Object-Centered & Text & \Checkmark \\
    PixelSynth~\cite{rockwell2021pixelsynth} & Outward-Facing & Single-Image & \Checkmark \\
    DiffDreamer~\cite{cai2023diffdreamer} & Outward-Facing & Single-Image & \Checkmark \\
    Xiang \emph{et al.}~\cite{xiang20233d} & Outward-Facing & Latent & \Checkmark \\
    CC3D~\cite{bahmani2023cc3d} & Outward-Facing & Layout & \Checkmark \\
    Text2Room~\cite{hollein2023text2room} & Outward-Facing & Text & \Checkmark \\
    Text2NeRF~\cite{zhang2023text2nerf} & Outward-Facing & Text & \Checkmark \\
\bottomrule
\end{tabular}
\label{tab:app_scene}
\end{table}

\subsection{General Scene Generation}
\label{sec:app:generic_gen}
Different from 3D human and face generation, which can use existing prior knowledge such as SMPL and 3DMM, general scene generation methods are more based on the similarity of semantics or categories to design a 3D model generation framework. Based on the differences in generation results, as shown in Fig.~\ref{fig:app_results_scene} and Tab.~\ref{tab:app_scene}, we further subdivide general scene generation into object-centered asset generation and outward-facing scene generation.

\subsubsection{Object-Centered Asset Generation}
The field of object-centered asset generation has seen significant advancements in recent years, with a focus on both textureless shape generation and textured asset generation. 
For the textureless shape generation, early works use GAN-based networks to learn a mapping from latent space to 3D object space based on specific categories of 3D data, such as 3D-GAN~\cite{wu2016learning}, HoloGAN~\cite{nguyen2019hologan}, and PlatonicGAN~\cite{henzler2019platonicgan}. 
However, limited by the generation capabilities of GANs, these methods can only generate rough 3D assets of specific categories. 
To improve the quality of generated results, SingleShapeGen~\cite{wu2022learning} leverages a pyramid of generators to generate 3D assets in a coarse to fine manner. Given the remarkable achievements of diffusion models in image generation, researchers are directing their attention towards the application of diffusion extensions in the realm of 3D generation.
Thus, subsequent methods \cite{luo2021diffusion, zhou20213d, hui2022neural, shue20233d, erkocc2023hyperdiffusion} explore the use of diffusion processes for 3D shape generation from random noise. 
In addition to these latent-based methods, another important research direction is text-driven 3D asset generation \cite{chen2019text2shape, liu2022towards}. 
For example, 3D-LDM \cite{nam20223d}, SDFusion \cite{cheng2023sdfusion}, and Diffusion-SDF \cite{li2023diffusion} achieve text-to-3D shape generation by designing the diffusion process in 3D feature space. 
Due to such methods requiring 3D datasets to train the diffusion-based 3D generators, they are limited to the training data in terms of the categories and diversity of generated results. 
By contrast, CLIP-Forge \cite{sanghi2022clip}, CLIP-Sculptor \cite{sanghi2023clip}, and Michelangelo \cite{zhao2023michelangelo} directly employ the prior of the pre-trained CLIP model to constrain the 3D generation process, effectively improving the generalization of the method and the diversity of generation results. 
Unlike the above latent-conditioned or text-driven 3D generation methods, to generate 3D assets with expected shapes, there are some works \cite{henzler2019escaping, chen2023deep3dsketch+} that explore image or sketch-conditioned generation.

In comparison to textureless 3D shape generation, textured 3D asset generation not only produces realistic geometric structures but also captures intricate texture details. For example, HoloGAN \cite{nguyen2019hologan}, GET3D~\cite{gao2022get3d}, and EG3D~\cite{chan2022efficient} employ GAN-based 3D generators conditioned on latent vectors to produce category-specific textured 3D assets. By contrast, text-driven 3D generation methods rely on the prior knowledge of pre-trained large-scale text-image models to enable category-free 3D asset generation. For instance, CLIP-Mesh~\cite{mohammad2022clip}, Dream Fields~\cite{jain2022zero}, and PureCLIPNeRF~\cite{lee2022understanding} employ the prior of CLIP model to constrain the optimization process and achieve text-driven 3D generation. Furthermore, DreamFusion~\cite{poole2022dreamfusion} and SJC~\cite{wang2023score} propose a score distillation sampling (SDS) method to achieve 3D constraint which priors extracted from pre-trained 2D diffusion models. Then, some methods further improve the SDS-based 3D generation process in terms of generation quality, multi-face problem, and optimization efficiency, such as Magic3D~\cite{lin2023magic3d}, Latent-NeRF~\cite{metzer2023latentnerf}, Fantasia3D~\cite{chen2023fantasia3d}, DreamBooth3D~\cite{raj2023dreambooth3d}, HiFA~\cite{zhu2023hifa}, ATT3D~\cite{lorraine2023att3d}, ProlificDreamer~\cite{wang2023prolificdreamer}, IT3D~\cite{chen2023it3d}, DreamGaussian~\cite{tang2023dreamgaussian}, and CAD~\cite{wan2023cad}. On the other hand, distinct from text-driven 3D generation, single-image-conditioned 3D generation is also a significant research direction \cite{liu2023zero, melas2023realfusion, chen2023single, wu2023hyperdreamer, kwak2023vivid }. 



\subsubsection{Outward-Facing Scene Generation}
Early scene generation methods often require specific scene data for training to obtain category-specific scene generators, such as GAUDI~\cite{bautista2022gaudi} and the work of Xiang \emph{et al.}~\cite{xiang20233d}, or implement a single scene reconstruction based on the input image, such as PixelSynth \cite{rockwell2021pixelsynth} and Worldsheet \cite{hu2021worldsheet}. However, these methods are either limited by the quality of the generation or by the extensibility of the scene. With the rise of diffusion models in image inpainting, various methods are beginning to use the scene completion capabilities of diffusion models to implement scene generation tasks \cite{cai2023diffdreamer, hollein2023text2room, zhang2023text2nerf}.
Recently, SceneScape \cite{fridman2023scenescape}, Text2Room \cite{hollein2023text2room}, Text2NeRF \cite{zhang2023text2nerf}, and LucidDreamer \cite{chung2023luciddreamer} propose progressive inpainting and updating strategies for generating realistic 3D scenes using pre-trained diffusion models. SceneScape and Text2Room utilize explicit polygon meshes as their 3D representation during the generation procedure. However, this choice of representation imposes limitations on the generation of outdoor scenes, resulting in stretched geometry and blurry artifacts in the fusion regions of mesh faces. In contrast, Text2NeRF and LucidDreamer adopt implicit representations, which offer the ability to model fine-grained geometry and textures without specific scene requirements. Consequently, Text2NeRF and LucidDreamer can generate both indoor and outdoor scenes with high fidelity.

\subsection{3D Editing}
\label{sec:app:editing}

Based on the region where editing happens, we classify the existing works into global editing and local editing.

\begin{figure}[t]
\begin{center}
   \includegraphics[width=1.0\linewidth]{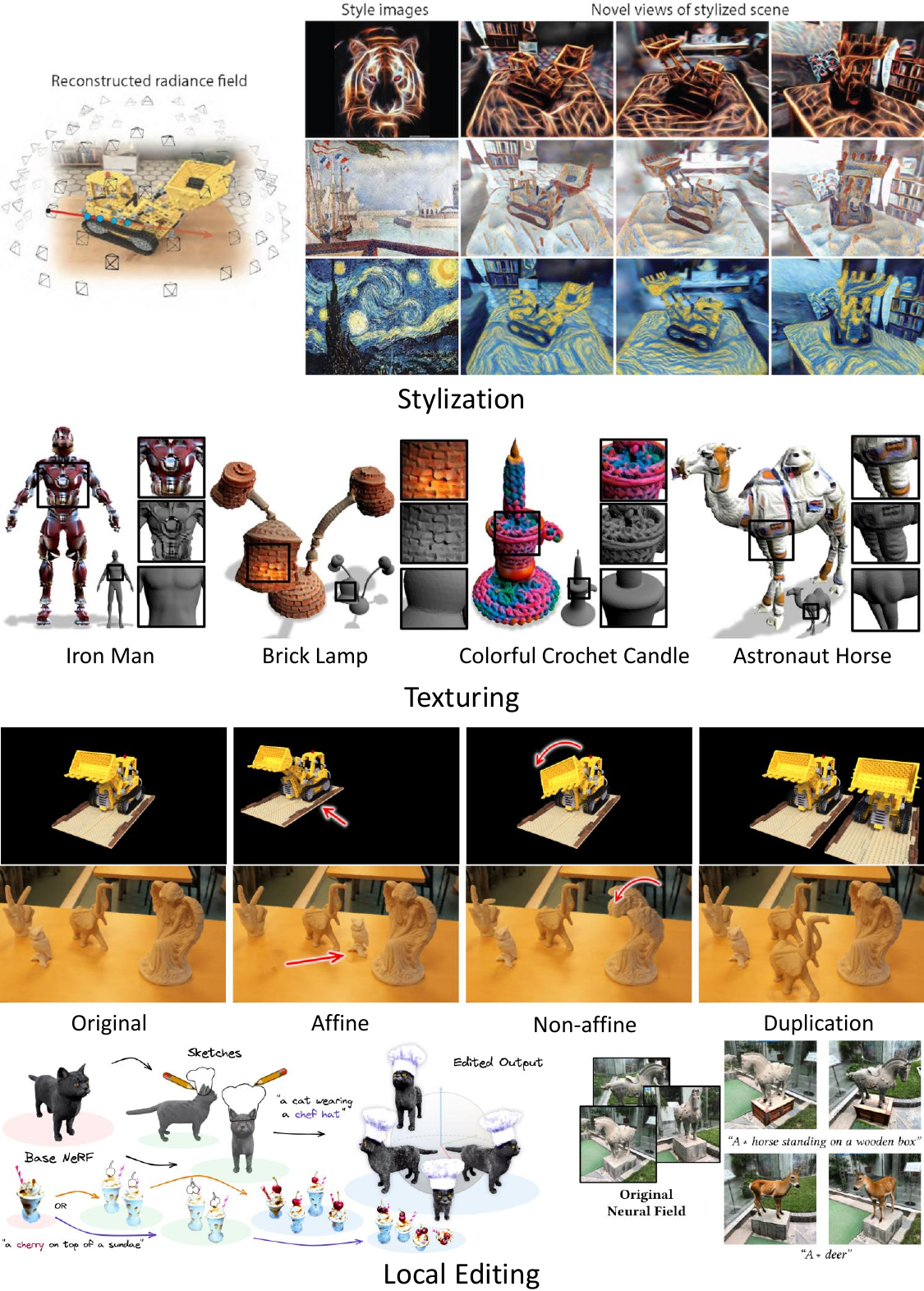}
\end{center}
\caption{
Representative 3D editing tasks.
Images adapted from ARF~\cite{zhang2022arf}, Text2Mesh~\cite{michel2022text2mesh}, NeRFShop~\cite{jambon2023nerfshop}, SKED~\cite{mikaeili2023sked}, and DreamEditor~\cite{zhuang2023dreameditor}.
}
\label{fig:app_face_gen}
\end{figure}

\subsubsection{Global Editing}
Global editing works aim at changing the appearance or geometry of the competing 3D scene globally. Different from local editing, they usually do not intentionally isolate a specific region from a complete and complicated scene or object.
Most commonly, they only care if the resultant scene is in a desired new ``style'' and resembles (maintains some features of) the original scene. Most representative tasks falling into this category include stylization~\cite{huang2021learning,huang2022stylizednerf,fan2022unified,zhang2022arf,wang2023nerfart,haque2023instruct}, 
and single-object manipulation (e.g. re-texturing~\cite{michel2022text2mesh,lei2022tango,metzer2023latentnerf,chen2023fantasia3d}) as shown in Fig.~\ref{fig:app_face_gen}.

\vspace{1mm}
\noindent\textbf{Stylization.} 
Early 3D scene stylization methods~\cite{huang2021learning,huang2022stylizednerf,fan2022unified,zhang2022arf} usually require style images to provide style reference. The 3D scene is optimized either in the style feature space using a Gram matrix~\cite{gatys2016image} or nearest neighbor feature matching~\cite{zhang2022arf} loss or in the image space using the output color of a deep image style transfer network~\cite{huang2017arbitrary}.
Later methods~\cite{wang2023nerfart,haque2023instruct} can support textual format style definition by utilizing the learned prior knowledge from large-scale language-vision models such as CLIP~\cite{radford2021clip} and Stable Diffusion~\cite{rombach2022ldm}. Other than commonly seen artistic style transfer, there also exist some special types of ``style'' manipulation tasks such as seasonal and illumination manipulations~\cite{li2022climatenerf,chen2022hallucinated,haque2023instruct,chen2022upst} and climate changes.

\vspace{1mm}
\noindent\textbf{Single-Object Manipulation.}
There are many papers specifically aim at manipulating a single 3D object.
For example, one representative task is texturing or painting a given 3D object (usually in mesh format)~\cite{michel2022text2mesh,lei2022tango,metzer2023latentnerf,chen2023fantasia3d,chen2023text2tex}. 
Except for diffuse albedo color and vertex displacement~\cite{michel2022text2mesh,ma2023x,liao2023tada}, other common property maps may be involved, including normal map~\cite{chen2023fantasia3d,lei2022tango}, roughness map~\cite{chen2023fantasia3d,lei2022tango}, specular map~\cite{lei2022tango}, and metallic map~\cite{chen2023fantasia3d}, etc.
A more general setting would be directly manipulating a NeRF-like object~\cite{wang2022clipnerf,lei2022tango,tseng2022cla,yang2022neumesh}.
Notably, the human face/head is one special type of object that has drawn a lot of interest~\cite{aneja2023clipface,zhang2023dreamface}.
In the meanwhile, many works focus on fine-grained local face manipulation, including expression and appearance manipulation~\cite{sun2022fenerf,sun2022ide,lin2023sketchfacenerf,jiang2022nerffaceediting,wu2022anifacegan,xu2023omniavatar, ma2022neural, zhang2022fdnerf} and face swapping~\cite{li20233d} since human face related understanding tasks (e.g. recognition, parsing, attribute classification) have been extensively studied previously.

\subsubsection{Local Editing}

Local editing tasks intentionally modify only a specific region, either manually provided (\cite{mikaeili2023sked,li2023focaldreamer,cheng2023progressive3d}) or automatically determined (\cite{yang2021learning,wu2022object,wu2023objectsdf++,kobayashi2022decomposing,jambon2023nerfshop}), of the complete scene or object.
Common local editing types include appearance manipulation~\cite{yang2022neumesh,zhuang2023dreameditor}, geometry deformation~\cite{jambon2023nerfshop,peng2022cagenerf,yuan2022nerf,tseng2022cla}, object-/semantic-level duplication/deletion and moving/removing~\cite{yang2021learning,wu2022object,kobayashi2022decomposing,wu2023objectsdf++}.
For example, NeuMesh~\cite{yang2022neumesh} supports several kinds of texture manipulation including swapping, filling, and painting since they distill a NeRF scene into a mesh-based neural representation.
NeRFShop~\cite{jambon2023nerfshop} and CageNeRF~\cite{peng2022cagenerf} transform/deform the volume bounded by a mesh cage, resulting in moved or deformed/articulated object.
SINE~\cite{bao2023sine} updates both the NeRF geometry and the appearance with geometry prior and semantic (image feature) texture prior as regularizations.

Another line of works (e.g. ObjectNeRF~\cite{yang2021learning}, ObjectSDF~\cite{wu2022object}, DFF~\cite{kobayashi2022decomposing}) focus on automatically decomposing the scene into individual objects or semantic parts during reconstruction, which is made possible by utilizing extra 2D image understanding networks (e.g. instance segmentation), and support subsequent object-level manipulations such as re-coloring, removal, displacement, duplication.

Recently, it is possible to create new textures and/or content only according to text description in the existing 3D scenes due to the success of large-scale text-to-image models (e.g. Stable Diffusion~\cite{rombach2022ldm}).
For example, instruct-NeRF2NeRF~\cite{haque2023instruct} iteratively updates the reference dataset images modified by a dedicated diffusion model~\cite{brooks2023instructpix2pix} and the NeRF model.
DreamEditor~\cite{zhuang2023dreameditor} performs local updates on the region located by text attention guided by score distillation sampling~\cite{poole2022dreamfusion}.
FocalDreamer~\cite{li2023focaldreamer} creates new geometries (objects) in the specified empty spaces according to the text input.
SKED~\cite{mikaeili2023sked} supports both creating new objects and modifying the existing part located in the region specified by the provided multi-view sketches.
\section{Open Challenges}\label{sec:open}
The quality and diversity of 3D generation results have experienced significant progress due to advancements in generative models, 3D representations, and algorithmic paradigms. Considerable attention has been drawn to 3D generation recently as a result of the success achieved by large-scale models in natural language processing and image generation. However, numerous challenges remain before the generated 3D models can meet the high industrial standards required for video games, movies, or immersive digital content in VR/AR. In this section, we will explore some of the open challenges and potential future directions in this field.

\noindent\textbf{Evaluation.}
Quantifying the quality of generated 3D models objectively is an important and not widely explored problem. Using metrics such as PSNR, SSIM, and F-Score to evaluate rendering and reconstruction results requires ground truth data on the one hand, but on the other hand, it can not comprehensively reflect the quality and diversity of the generated content. In addition, user studies are usually time-consuming, and the study results tend to be influenced by the bias and number of surveyed users. Metrics that capture both the quality and diversity of the results like FID can be applied to 3D data, but may not be always aligned with 3D domain and human preferences. Better metrics to judge the results objectively in terms of generation quality, diversity, and matching degree with the conditions still need further exploration.

\noindent\textbf{Dataset.}
Unlike language or 2D image data which can be easily captured and collected, 3D assets often require 3D artists or designers to spend a significant amount of time using professional software to create. Moreover, due to the different usage scenarios and creators' personal styles, these 3D assets may differ greatly in scale, quality, and style, increasing the complexity of 3D data. Specific rules are needed to normalize this diverse 3D data, making it more suitable for generation methods. A large-scale, high-quality 3D dataset is still highly desirable in 3D generation. Meanwhile, exploring how to utilize extensive 2D data for 3D generation could also be a potential solution to address the scarcity of 3D data.

\noindent\textbf{Representation.}
Representation is an essential part of the 3D generation, as we discuss various representations and the associated methods in Sec. \ref{sec:representation}. Implicit representation is able to model complex geometric topology efficiently but faces challenges with slow optimization; explicit representation facilitates rapid optimization convergence but struggles to encapsulate complex topology and demands substantial storage resources; Hybrid representation attempts to consider the trade-off between these two, but there are still shortcomings. In general, we are motivated to develop a representation that balances optimization efficiency, geometric topology flexibility, and resource usage.

\noindent\textbf{Controllability.}
The purpose of the 3D generation technique is to generate a large amount of user-friendly, high-quality, and diverse 3D content in a cheap and controllable way. However, embedding the generated 3D content into practical applications remains a challenge: most methods~\cite{poole2022dreamfusion, chan2022efficient, yang2019pointflow} rely on volume rendering or neural rendering, and fail to generate content suitable for rasterization graphics pipeline. As for methods~\cite{chen2023fantasia3d, wang2023prolificdreamer, tang2023dreamgaussian} that generate the content represented by polygons, they do not take into account layout (\eg the rectangular plane of a table can be represented by two triangles) and high-quality UV unwrapping and the generated textures also face some issues such as baked shadows. These problems make the generated content unfavorable for artist-friendly interaction and editing. Furthermore, the style of generated content is still limited by training datasets.
Furthermore, the establishment of comprehensive toolchains is a crucial aspect of the practical implementation of 3D generation. In modern workflows, artists use tools (\eg LookDev) to harmonize 3D content by examining and contrasting the relighting results of their materials across various lighting conditions. Concurrently, modern Digital Content Creation (DCC) software offers extensive and fine-grained content editing capabilities. It is promising to unify 3D content produced through diverse methods and establish tool chains that encompass abundant editing capabilities.

\noindent\textbf{Large-scale Model.}
Recently, the popularity of large-scale models has gradually affected the field of 3D generation. Researchers are no longer satisfied with using distillation scores that use large-scale image models as the priors to optimize 3D content, but directly train large-scale 3D models. MeshGPT~\cite{siddiqui2023meshgpt} follows large language models and adopts a sequence-based approach to autoregressively generate sequences of triangles in the generated mesh. MeshGPT takes into account layout information and generates compact and sharp meshes that match the style created by artists. Since MeshGPT is a decoder-only transformer, compared with the optimization-based generation, it gets rid of inefficient multi-step sequential optimization, achieving rapid generation.
Despite this, MeshGPT's performance is still limited by training datasets and can only generate regular furniture objects. But there is no doubt that large-scale 3D generation models have great potential worth exploring.

\section{Conclusion}\label{sec:conclusion}
In this work, we present a comprehensive survey on 3D generation, encompassing four main aspects: 3D representations, generation methods, datasets, and various applications. We begin by introducing the 3D representation, which serves as the backbone and determines the characteristics of the generated results. Next, we summarize and categorize a wide range of generation methods, creating an evolutionary tree to visualize their branches and developments. Finally, we provide an overview of related datasets, applications, and open challenges in this field. The realm of 3D generation is currently witnessing explosive growth and development, with new work emerging every week or even daily. We hope this survey offers a systematic summary that could inspire subsequent work for interested readers.



\bibliographystyle{IEEEtran}
\bibliography{eigbib}




\end{document}